\documentclass{article} 
\usepackage{graphicx}  
\usepackage{float}
\usepackage{amsmath}
\usepackage{amssymb}
\usepackage{amsthm}
\usepackage{url}

\usepackage{authblk}

\floatstyle{ruled}
\newfloat{algorithm}{tbp}{loa}
\providecommand{\algorithmname}{Algorithm}
\floatname{algorithm}{\protect\algorithmname}

\theoremstyle{plain}
\newtheorem{thm}{\protect\theoremname}
\theoremstyle{definition}
\newtheorem{defn}[thm]{\protect\definitionname}
\theoremstyle{plain}
\newtheorem{lem}[thm]{\protect\lemmaname}
\theoremstyle{plain}
\newtheorem{prop}[thm]{\protect\propositionname}

\ifx\proof\undefined
\newenvironment{proof}[1][\protect\proofname]{\par
	\normalfont\topsep6\p@\@plus6\p@\relax
	\trivlist
	\itemindent\parindent
	\item[\hskip\labelsep\scshape #1]\ignorespaces
}{%
	\endtrivlist\@endpefalse
}
\providecommand{\proofname}{Proof}
\fi

\usepackage{algorithmic}

\providecommand{\corollaryname}{Corollary}
\providecommand{\definitionname}{Definition}
\providecommand{\lemmaname}{Lemma}
\providecommand{\theoremname}{Theorem}
\providecommand{\propositionname}{Proposition}
\newcommand{\citep}[1]{\cite{#1}}
\newcommand{\citet}[1]{\cite{#1}}

\begin{document}
%
\title{Which Factorization Machine Modeling is Better:\\A Theoretical Answer with Optimal Guarantee}
\author[1]{Ming Lin\thanks{Accepted by The Thirty-Third AAAI Conference on Artificial Intelligence (AAAI-19).}}
\author[2]{Shuang Qiu}
\author[2]{Jieping Ye}
\author[1]{Xiaomin Song}
\author[1]{Qi Qian}
\author[1]{Liang Sun}
\author[1]{Shenghuo Zhu}
\author[1]{Rong Jin}
\affil[1]{Alibaba Group, Bellevue, USA. \protect\\
\{ming.l, xiaomin.song, qi.qian,liang.sun, shenghuo.zhu, jinrong.jr\}@alibaba-inc.com}
\affil[2]{Department of Computational Medicine and Bioinformatics, University of Michigan, Ann Arbor, USA. \protect\\
\{qiush, jpye\}@umich.edu}
\date{January 27th, 2019}

%

\maketitle

\begin{abstract}
Factorization machine (FM) is a popular machine learning model to
capture the second order feature interactions. The optimal learning
guarantee of FM and its generalized version is not yet developed.
For a rank $k$ generalized FM of $d$ dimensional input, the previous
best known sampling complexity is $\mathcal{O}[k^{3}d\cdot\mathrm{polylog}(kd)]$
under Gaussian distribution. This bound is sub-optimal comparing to
the information theoretical lower bound $\mathcal{O}(kd)$. In this
work, we aim to tighten this bound towards optimal and generalize
the analysis to sub-gaussian distribution. We prove that when the
input data satisfies the so-called $\tau$-Moment Invertible Property,
the sampling complexity of generalized FM can be improved to $\mathcal{O}[k^{2}d\cdot\mathrm{polylog}(kd)/\tau^{2}]$.
When the second order self-interaction terms are excluded in the generalized
FM, the bound can be improved to the optimal $\mathcal{O}[kd\cdot\mathrm{polylog}(kd)]$
up to the logarithmic factors. Our analysis also suggests that
the positive semi-definite constraint in the conventional FM is redundant
as it does not improve the sampling complexity while making the model
difficult to optimize. We evaluate our improved FM model in real-time high precision
GPS signal calibration task to validate its superiority.
\end{abstract}

\section{Introduction}

Factorization machine (FM) \citep{rendle_factorization_2010,Bayer:2017:GCD:3038912.3052694,Juan:2017:FFM:3041021.3054185,juan_field-aware_2016,Zhao:2017:MBR:3097983.3098063,yamada2017convex,luo2018sketched,lin2018online}
is a popular linear regression model to capture the second order feature
interactions. It has been found effective in various applications,
including recommendation systems \citep{rendle_factorization_2010}
, CTR prediction \citep{juan_field-aware_2016}, computational medicine
\citep{lin_multi-task_2016} , social network \citep{hong_co-factorization_2013}
and so on. Intuitively speaking, the second order feature interactions
consider the factors jointly affecting the output. On the theoretical
side, FM is closely related to the symmetric matrix sensing \citep{kueng_low_2014,cai_rop:_2015,yurtsever_sketchy_2017}
and phase retrieval \citep{candes_phase_2011}.
While the conventional FM only considers the second order feature
interactions, it is possible to extend the conventional FM to the
high order functional space which leads to the Polynomial Network
model \citep{blondel_polynomial_2016}. FM is the cornerstone in modern
machine learning research as it abridges the linear
regression and high order polynomial regression. It is therefore important to
understand the theoretical foundation of FM.

Given an instance $\boldsymbol{x}\in\mathbb{R}^{d}$, the conventional
FM assumes that the label $y\in\mathbb{R}$ of $\boldsymbol{x}$ is
generated by
\begin{align}
 & y=\boldsymbol{x}{}^{\top}\boldsymbol{w}^{*}+\boldsymbol{x}{}^{\top}M^{*}\boldsymbol{x}\quad\mathrm{rank}(M^{*})\leq k\label{eq:FM-model}
\end{align}
where $\{\boldsymbol{w}^{*},M^{*}\}$ are the first order and the
second order coefficients respectively. In the original FM paper \citep{rendle_factorization_2010},
the authors additionally assumed that $M^{*}$ is generated from a
low-rank positive semi-definite (PSD) matrix with all its diagonal
elements subtracted. That is
\begin{equation}
M^{*}=U^{*}U^{*}{}^{\top}-\mathrm{diag}(U^{*}U^{*}{}^{\top})\ .\label{eq:FM-PSD-and-diag0-assumption}
\end{equation}
 Eq. (\ref{eq:FM-PSD-and-diag0-assumption}) consists of two parts.
We call the first part $U^{*}U{}^{*}{}^{\top}$ as the PSD constraint
and the second part $-\mathrm{diag}(\cdot)$ as the diagonal-zero
constraint. Our key question in this work is whether the FM model
(\ref{eq:FM-model}) can be learned by $\mathcal{O}[kd\log(kd)]$
observations and how the two additional constraints help the generalization
ability of FM.

Although the FM has been widely applied , there is little research
exploring the theoretical properties of the FM to answer the above
key question. A naive analysis directly following the sampling complexity
of the linear model would suggest $O(d^{2})$ samples to recover $\{\boldsymbol{w}^{*},M^{*}\}$
which is too loose. When $\boldsymbol{w}^{*}=0$ and $M^{*}$ is symmetric,
Eq. (\ref{eq:FM-model}) is equal to the symmetric matrix sensing
problem. \citet{cai_rop:_2015} proved the sampling complexity of
this special case on well-bounded sub-gaussian distribution using
trace norm convex programming under the $\ell_{2}/\ell_{1}$-RIP condition.
\citet{yurtsever_sketchy_2017} developed a conditional gradient descent
solver to recover $M^{*}$. However, when $\boldsymbol{w}^{*}\not=\boldsymbol{0}$
the above methods and the theoretical results are no longer applicable.
\citet{blondel_convex_2015} considered a convexified formulation
of FM. Their FM model requires solving a trace-norm regularized loss
function which is computationally expensive. They did not provide
statistical learning guarantees for the convexified FM.

To the best of our knowledge, the most recent research dealing with
the theoretical properties of the FM is \citep{ming_lin_non-convex_2016}.
In their study, the authors argued that the two constraints proposed
in \citep{rendle_factorization_2010} can be removed if $\boldsymbol{x}$
is sampled from the standard Gaussian distribution. However, their
analysis heavily relies on the rotation invariance of the Gaussian
distribution therefore cannot be generalized to non-Gaussian cases.
Even limiting on the Gaussian distribution, the sampling complexity
given by their analysis is $\mathcal{O}[k^{3}d\cdot\mathrm{polylog}(kd)]$
which is worse than the information-theoretic lower bound $\mathcal{O}(kd)$.
It is still an open question whether the FM can be learned with $\mathcal{O}(kd)$
samples and whether both constraints in the original formulation are
necessary to make the model learnable.

In this work, we answer the above questions affirmatively. We show
that when the data is sampled from sub-gaussian distribution and satisfies
the so-called $\tau$-Moment Invertible Property (MIP), the generalized
FM (without constraints) can be learned by $\mathcal{O}[k^{2}d/\tau^{2}\cdot\mathrm{polylog}(kd)]$
samples. The PSD constraint is not necessary to achieve this sharp
bound. Actually the PSD constraint is harmful as it introduces asymmetric
bias on the value $y$ (see Experiment section). The optimal sampling
complexity $\mathcal{O}[kd\cdot\mathrm{polylog}(kd)]$ is achievable
if we further constrain that the diagonal elements of $M^{*}$ are
zero. This is not an artificial constraint but there is information-theoretic
limitation prevents us recovering the diagonal elements of $M^{*}$
on sub-gaussian distribution. Finally inspired by our theoretical
results, we propose an improved version of FM, called iFM, which removes
the PSD constraint and inherits the diagonal-zero constraint from
the conventional modeling. Unlike the generalized FM, the sampling
complexity of the iFM does not depend on the MIP constant of the data
distribution.

The remainder of this paper is organized as follows. We revisit the
modeling of the FM in Section 2 and show that the conventional modeling
is sub-optimal when considered in a more general framework. In Section
3 we present the learning guarantee of the generalized FM on sub-gaussian
distribution. We propose the high order moment elimination technique
to overcome a difficulty in our convergence analysis. Based on our
theoretical results, we propose the improved model iFM. Section 4
conducts numerical experiments on synthetic and real-world datasets
to validate the superiority of iFM over the conventional FM . Section
5 encloses this work.

\section{A Revisit of Factorization Machine }

In this section, we revisit the modeling design of the FM and its
variants. We briefly review the original formulation of the conventional
FM to raise several questions about its optimality. We then highlight
previous studies trying to establish the theoretical foundation of
the FM modeling. Based on the above survey, we motivate our study
and present our main results in the next section.

In their original paper, \citet{rendle_factorization_2010} assumes
that the feature interaction coefficients in the FM can be embedded
in a $k$-dimensional latent space. That is,
\begin{align}
 & y=\sum_{i=1}^{d}w_{i}x_{i}+\sum_{i=1}^{d}\sum_{j=i+1}^{d}\left\langle \boldsymbol{u}_{i},\boldsymbol{u}_{j}\right\rangle x_{i}x_{j}\label{eq:orig-FM-formulation}
\end{align}
 where $\boldsymbol{u}_{i}$ is a $k\times1$ vector. The original
formulation Eq. (\ref{eq:orig-FM-formulation}) is equivalent to Eq.
(\ref{eq:FM-model}) with constraint Eq. (\ref{eq:FM-PSD-and-diag0-assumption}).
While the low-rank assumption is standard in the matrix sensing literature,
the PSD and the diagonal-zero constraints are not. A critical question
is whether the two additional constraints are necessary or removable.
Indeed we have strong reasons to remove both constraints.

The reasons to remove the diagonal-zero constraint are straightforward.
First there is no theoretical result so far to motivate this constraint.
Secondly subtracting the diagonal elements will make the second order
derivative w.r.t. $U$ non-PSD. This will raise many technical difficulties
in optimization and learning theory as many research works assume
convexity in their analysis.

The PSD constraint in the original FM modeling is the second term
we wish to remove. Let us temporally forget about the diagonal-zero
constraint and focus on the PSD constraint only. Obviously relaxing
$UU^{\top}$ with $UV^{\top}$ will make the model more flexible.
A more serious problem of the PSD constraint is that it implicitly
assumes that the label $y$ is more likely to be ``positive''. This
will introduce asymmetric bias about the distribution of $y$. To
see this, suppose $\hat{M}=U^{*}U^{*}{}^{\top}=\bar{U}\Sigma\bar{U}{}^{\top}$
where $\bar{U}$ is the eigenvector matrix of $\hat{M}$. We call
$\bar{U}$ the second order feature mapping matrix induced by $\hat{M}$
since $\boldsymbol{x}{}^{\top}\hat{M}\boldsymbol{x}=(\bar{U}{}^{\top}\boldsymbol{x}){}^{\top}\Sigma(\bar{U}{}^{\top}\boldsymbol{x})$.
The eigenvalue matrix $\Sigma$ is the weights for the mapped features
$\bar{U}{}^{\top}\boldsymbol{x}$. As $\hat{M}$ is constrained to
be PSD, the weights of $\bar{U}{}^{\top}\boldsymbol{x}$ cannot be
negative. In other words, the PSD constraint prevents the model learning
patterns from negative class. Please check the Experiment section
for more concrete examples.

Another issue of the PSD constraint raised is the difficulty in optimization.
Suppose we choose least square as the loss function in FM. By enforcing
$\hat{M}=UU{}^{\top}$, the loss function is a fourth order polynomial
of $U$. This makes the initialization of $U$ difficult since the
scale of the initial $U^{(0)}$ will  affect the convergence rate.
Clearly we cannot initialize $U^{(0)}=0$ since the gradient w.r.t.
$U$ will be zero. On the other hand, we cannot initialize $U^{(0)}$
to have a large norm otherwise the problem will be ill-conditioned.
This is because the spectral norm of the second order derivative w.r.t.
$U$ will be proportional to $\|U^{(0)}\|_{2}^{2}$ therefore the
(local) condition number depends on $\|U^{(0)}\|_{2}$. In practice,
it is usually difficult to figure out the optimal scale of $\|U^{(0)}\|_{2}$
resulting vanishing or explosion gradient norms. If we decouple the
$U$ and $V$, we can initialize $\|U^{(0)}\|_{2}=1$ and $V^{(0)}=0$.
Then by alternating gradient descent, the decoupled FM model is easy
to optimize.

In summary, the theoretical foundation of the FM is still not well-developed.
On one hand, it is unclear whether the conventional FM modeling is
optimal and on the other hand, there is strong motivation to modify
the conventional formulation based on heuristic intuition. This inspires
our study of the optimal modeling of the FM driven by theoretical
analysis which is presented in the next section.

\section{Main Results}

In this section, we present our main results on the theoretical guarantees
of the FM and its improved version iFM. We first give a sharp complexity
bound for the generalized FM on sub-gaussian distribution. We show
that the recovery error of the diagonal elements of $M^{*}$ depends
on a so-called $\tau$-MIP condition of the data distribution. The
sampling complexity bound can be improved to optimal by the diagonal-zero
constraint.

We introduce a few more notations needed in this section. Suppose
$\boldsymbol{x}$ is sampled from coordinate sub-gaussian with zero
mean and unit variance. The element-wise third order moment of $\boldsymbol{x}$
is denoted as $\boldsymbol{\kappa}^{*}\triangleq\mathbb{E}\boldsymbol{x}^{3}$
and the fourth order moment is $\boldsymbol{\phi}^{*}\triangleq\mathbb{E}\boldsymbol{x}^{4}$.
All training instances are sampled identically and independently (i.i.d.).
Denote the feature matrix $X=[\boldsymbol{x}^{(1)},\cdots,\boldsymbol{x}^{(n)}]\in\mathbb{R}^{d\times n}$
and the label vector $\boldsymbol{y}=[y_{1},\cdots,y_{n}]{}^{\top}\in\mathbb{R}^{n}$.
$\mathcal{D}(\cdot)$ denotes the diagonal function. For any two matrices
$A$ and $B$, we denote their Hadamard product as $A\circ B$. The
element-wise squared matrix is defined by $A^{2}\triangleq A\circ A$.
For a non-negative real number $\xi\geq0$, the symbol $O(\xi)$ denotes
some perturbation matrix whose spectral norm is upper bounded by $\xi$
. The $i$-th largest singular value of matrix $M$ is $\sigma_{i}(M)$
. We abbreviate $\sigma_{i}^{*}\triangleq\sigma_{i}(M^{*})$. To abbreviate
our high probability bounds, given a probability $\eta$, we use the
symbol $C_{\eta}$ and $c_{\eta}$ to denote some polynomial logarithmic
factors in $1/\eta$ and any other necessary variables that do not
change the polynomial order of the upper bounds.

\subsection{Limitation of The Generalized FM}

In order to derive the theoretical optimal FM models, we begin with
the most general formulation of FM, that is, with no constraint except
low-rank:
\begin{align}
 & y=\boldsymbol{x}{}^{\top}\boldsymbol{w}^{*}+\boldsymbol{x}{}^{\top}M^{*}\boldsymbol{x}\quad\mathrm{s.t.}\ M^{*}=U^{*}V^{*}{}^{\top}\ .\label{eq:gFM-model}
\end{align}
 Clearly $M^{*}$ must be symmetric but for now this does not matter.
Eq. (4) is called the generalized FM \citep{ming_lin_non-convex_2016}.
It is proved that when $\boldsymbol{x}$ is sampled from the Gaussian
distribution, Eq. (\ref{eq:gFM-model}) can be learned by $\mathcal{O}(k^{3}d)$
training samples. Although this bound is not optimal, \citet{ming_lin_non-convex_2016}
showed the possibility to remove Eq. (\ref{eq:FM-PSD-and-diag0-assumption})
on the Gaussian distribution. However, their result no longer holds
true on non-Gaussian distributions. In the following, we will show
that the learning guarantee for the generalized FM on sub-gaussian
distribution is much more complex than the Gaussian one.

Our first important observation is that model (\ref{eq:gFM-model})
is not always learnable on all sub-gaussian distributions.
\begin{prop}
\label{prop:gFM-not-learnable-on-bernoulli distribution} When $\boldsymbol{x}\in\{-1,+1\}^{d}$,
the generalized FM is not learnable.
\end{prop}

The above observation is easy to verify since $\boldsymbol{x}{}^{\top}M^{*}\boldsymbol{x}=\mathrm{tr}(M^{*})$
when $\boldsymbol{x}\in\{-1,+1\}^{d}$. Therefore at least the diagonal
elements of $M^{*}$ cannot be recovered at all. Proposition \ref{prop:gFM-not-learnable-on-bernoulli distribution}
shows that there is information-theoretic limitation to learn the
generalized FM on sub-gaussian distribution. In our analysis, we find
that such limitation is related to a property of the data distribution
which we call the \emph{Moment Invertible Property } (MIP).
\begin{defn}[Moment Invertible Property]
 \label{def:Moment-Invertible-Property} A zero mean unit variance
sub-gaussian distribution $\mathbb{P}(x)$ is called $\tau$-Moment
Invertible if $|\phi-1-\kappa^{2}|\geq\tau$ for some constants $\tau\geq0$,
$\phi\triangleq\mathbb{E}x^{4}$, $\kappa\triangleq\mathbb{E}x^{3}$.

With the MIP condition, the following theorem shows that the generalized
FM is learnable via alternating gradient descent.
\end{defn}

\begin{thm}
\label{thm:global-convergence-rate-of-Moment-Estimation-Sequence}
Suppose  $\boldsymbol{x}$ is sampled from a $\tau$-MIP sub-gaussian
distribution with $\boldsymbol{y}$ generated by Eq. (\ref{eq:gFM-model}).
Then with probability at least $1-\eta$, there is an alternating
gradient descent based method which can achieve the recovery error
after $t$ iteration such that
\begin{align*}
 & \|\boldsymbol{w}^{(t)}-\boldsymbol{w}^{*}\|_{2}+\|M^{(t)}-M^{*}\|_{2}\leq\\
 & \quad[(2\sqrt{5}\sigma_{1}^{*}/\sigma_{k}^{*}+2)\delta]^{t}(\|\boldsymbol{w}^{*}\|_{2}+\|M^{*}\|_{2})\ ,
\end{align*}
 provided
\begin{align*}
 & n\geq\frac{C_{\eta}}{\delta^{2}}(p+1)^{2}\max\{p\tau^{-2},(\sqrt{k}+|\mathrm{tr}(M)|/\|M\|_{2})^{2}d\}\\
 & p\triangleq\max\{1,\|\boldsymbol{\kappa}^{*}\|_{\infty},\|\boldsymbol{\phi}^{*}-3\|_{\infty},\|\boldsymbol{\phi}^{*}-1\|_{\infty}\}\\
 & \delta\leq\min\{\frac{1}{2\sqrt{5}\sigma_{1}^{*}/\sigma_{k}^{*}+2},\frac{1}{2\sqrt{5}}\sigma_{k}^{*}[\|\boldsymbol{w}^{*}\|_{2}+\|M^{*}\|_{2}]^{-1}\}\ .
\end{align*}
\end{thm}

Theorem \ref{thm:global-convergence-rate-of-Moment-Estimation-Sequence}
is our key result. In Theorem \ref{thm:global-convergence-rate-of-Moment-Estimation-Sequence},
we measure the quality of our estimation by the recovery error
\[
\epsilon_{t}\triangleq\|\boldsymbol{w}^{(t)}-\boldsymbol{w}^{*}\|_{2}+\|M^{(t)}-M^{*}\|_{2}\ .
\]
 The recovery error decreases linearly along the steps of alternating
iteration with rate $\delta\approx\mathcal{O}(1/\sqrt{n})$. The sampling
complexity is on order of $\max\{\mathcal{O}(k^{2}d\},\mathcal{O}(1/\tau^{2})\}$.
This bound delivers two messages. First when the distribution is close
to the Gaussian distribution, $\tau\approx2$ and the bound is controlled
by $\mathcal{O}(k^{2}d)$. This result improves the previous $\mathcal{O}(k^{3}d)$
given by \citet{ming_lin_non-convex_2016} for the Gaussian distribution.
Secondly when $\tau$ is small, the sampling complexity is proportional
to $\mathcal{O}(1/\tau^{2})$. The sampling complexity will even trend
to infinite when the data follows the binary Bernoulli distribution
where $\tau=0$. Therefore the $\tau$-MIP condition provides a sufficient
condition to make the generalized FM learnable.

We have not given any detail about the alternating gradient descent
algorithm mentioned in Theorem \ref{thm:global-convergence-rate-of-Moment-Estimation-Sequence}.
We find that it is difficult to prove the convergence rate following
the conventional alternating gradient descent framework. To address
this difficulty, we use a high order moment elimination technique
in the next subsection in the convergent analysis.

\subsection{Alternating Gradient Descent with High Order Moment Elimination}

\begin{algorithm}
\begin{algorithmic}[1]

\REQUIRE The mini-batch size $n$; number of total update $T$; training
instances $X^{(t)}\triangleq[\boldsymbol{x}^{(t,1)},\boldsymbol{x}^{(t,2)},\cdots,\boldsymbol{x}^{(t,n)}]$,
$\boldsymbol{y}^{(t)}\triangleq[y^{(t,1)},y^{(t,2)},\cdots,y^{(t,n)}]{}^{\top}$;
rank $k\geq1$ .

\ENSURE $\boldsymbol{w}^{(T)},U^{(T)},V^{(T)},M^{(t)}\triangleq U^{(t)}V^{(t)}{}^{\top}$.

\STATE Retrieve $n$ training instances to estimate the third and
fourth order moments $\boldsymbol{\kappa}$ and $\boldsymbol{\phi}$
.

\STATE Compute $G$ and $H$ in Eq. (\ref{eq:G}) and (\ref{eq:H}).

\STATE Initialize $\boldsymbol{w}^{(0)}=\boldsymbol{0}$, $V^{(0)}=0$.
$\bar{U}^{(0)}=\mathrm{SVD}(\mathcal{M}^{(0)}(\boldsymbol{y}^{(0)}),k)$,
that is the top-$k$ singular vectors.

\FOR{$t=1,2,\cdots,T$}

\STATE Retrieve $n$ training instances $X^{(t)},\boldsymbol{y}^{(t)}$
, compute $\mathcal{M}^{(t)}$ in Eq. (\ref{eq:mathcal-M-t}) and
update $U^{(t)}$ and $V^{(t)}$ as:
\begin{align*}
 & \hat{\boldsymbol{y}}^{(t)}=X^{(t)}{}^{\top}\boldsymbol{w}^{(t-1)}+\mathcal{A}^{(t)}(U^{(t-1)}V^{(t-1)}{}^{\top})\\
 & U^{(t)}=V^{(t-1)}-\mathcal{M}^{(t)}(\hat{\boldsymbol{y}}^{(t)}-\boldsymbol{y}^{(t)})\bar{U}^{(t-1)}\\
 & \{\bar{U}^{(t)},R^{(t)}\}=\mathrm{QR}(U^{(t)})\\
 & V^{(t)}=V^{(t-1)}\bar{U}^{(t-1)}{}^{\top}\bar{U}^{(t)}-\mathcal{M}^{(t)}(\hat{\boldsymbol{y}}^{(t)}-\boldsymbol{y}^{(t)})\bar{U}^{(t)}
\end{align*}

\STATE Compute $\mathcal{W}^{(t)}$ in Eq. (\ref{eq:mathcal-W-t})
and update $\boldsymbol{w}^{(t)}=\boldsymbol{w}^{(t-1)}-\mathcal{W}^{(t)}(\hat{\boldsymbol{y}}^{(t)}-\boldsymbol{y}^{(t)})\ .$

\ENDFOR

\STATE \textbf{Output:} $\boldsymbol{w}^{(T)},\bar{U}^{(T)},V^{(T)}$
.\textbf{ }

\end{algorithmic}

\caption{Alternating Gradient Descent with High Order Moment Elimination}

\label{alg:moment-estimation-sequence-method}
\end{algorithm}

In this subsection, we will construct an alternating gradient descent
algorithm which achieves the convergence rate and the sampling complexity
in Theorem \ref{thm:global-convergence-rate-of-Moment-Estimation-Sequence}.
We first show that the conventional alternating gradient descent cannot
be applied directly to prove Theorem \ref{thm:global-convergence-rate-of-Moment-Estimation-Sequence}.
Then a high order moment elimination technique is proposed to overcome
the difficulty.

The generalized FM defined in Eq. (\ref{eq:gFM-model}) can be written
in the matrix form
\begin{equation}
\boldsymbol{y}=X{}^{\top}\boldsymbol{w}^{*}+\mathcal{A}(M^{*})\label{eq:second-order-linear-model-matrix-form}
\end{equation}
 where the operator $\mathcal{A}(\cdot):\mathbb{R}^{d\times d}\rightarrow\mathbb{R}^{d}$
is defined by $\mathcal{A}(M)\triangleq[\boldsymbol{x}^{(1)}{}^{\top}M\boldsymbol{x}^{(1)},\cdots,\boldsymbol{x}^{(n)}{}^{\top}M\boldsymbol{x}^{(n)}]$.
The adjoint operator of $\mathcal{A}$ is $\mathcal{A}'$. To recover
$\{\boldsymbol{w}^{*},M^{*}\}$, we minimize the square loss function
\begin{equation}
\min_{\boldsymbol{w},U,V}\ \mathcal{L}(\boldsymbol{w},U,V)\triangleq\frac{1}{2n}\|X{}^{\top}\boldsymbol{w}+\mathcal{A}(UV{}^{\top})-\boldsymbol{y}\|^{2}\ .\label{eq:non-convex-optimization-loss}
\end{equation}

A straightforward idea to prove Theorem \ref{thm:global-convergence-rate-of-Moment-Estimation-Sequence}
is to show that the alternating gradient descent will converge. However,
we find that this is difficult in our problem. To see this, let us
compute the expected gradient of $\mathcal{L}(\boldsymbol{w}^{(t)},U^{(t)},V^{(t)})$
with respect to $V^{(t)}$ at step $t$.
\begin{align*}
\mathbb{E}\nabla_{V}\mathcal{L}(\boldsymbol{w}^{(t)},U^{(t)},V^{(t)})= & 2(M^{(t)}-M^{*})U^{(t)}+F^{(t)}U^{(t)}
\end{align*}
 where
\begin{align*}
F^{(t)} & \triangleq\mathrm{tr}(M^{(t)}-M^{*})I+\mathcal{D}(\boldsymbol{\phi}-3)\mathcal{D}(M^{(t)}-M^{*})\\
 & +\mathcal{D}(\boldsymbol{\kappa})\mathcal{D}(\boldsymbol{w}^{(t)}-\boldsymbol{w}^{*})\ .
\end{align*}
In previous studies, one expects $\mathbb{E}\nabla\mathcal{L}\approx I$.
However, this is no longer the case in our problem. Clearly $\|\frac{1}{2}\mathbb{E}\nabla\mathcal{L}-I\|_{2}$
is dominated by $\|\boldsymbol{\kappa}\|_{\infty}$ and $\|\boldsymbol{\phi}-3\|_{\infty}$
. For non-Gaussian distributions, these two perturbation terms could
dominate the gradient norm. Similarly the gradient of $\boldsymbol{w}$
is biased by $O(\|\boldsymbol{\kappa}\|_{\infty})$.

The difficulty to follow the conventional gradient descent analysis
inspires us to look for a new convergence analysis technique. The
perturbation term $F^{(t)}$ consists of high order moments of the
sub-gaussian variable $\boldsymbol{x}$. It might be possible to construct
a sequence of another high order moments to eliminate these perturbation
terms. We call this idea the high order moment elimination method.
The next question is whether the desired moments exist and how to
construct them efficiently. Unfortunately, this is impossible in general.
A sufficient condition to ensure the existence of the elimination
sequence is that the data distribution satisfies the $\tau$-MIP condition.

To construction an elimination sequence, for any $\boldsymbol{z}\in\mathbb{R}^{n}$
and $M\in\mathbb{R}^{d\times d}$, define functions
\begin{align*}
 & \mathcal{P}^{(t,0)}(\boldsymbol{z})\triangleq\boldsymbol{1}{}^{\top}\boldsymbol{z}/n,\quad\mathcal{P}^{(t,1)}(\boldsymbol{z})\triangleq X^{(t)}\boldsymbol{z}/n\\
 & \mathcal{P}^{(t,2)}(\boldsymbol{z})\triangleq(X^{(t)})^{2}\boldsymbol{z}/n-\mathcal{P}^{(t,0)}(\boldsymbol{z})\\
 & \mathcal{A}^{(t)}(M)\triangleq\mathcal{D}(X^{(t)}{}^{\top}MX^{(t)})\\
 & \mathcal{H}^{(t)}(\boldsymbol{z})\triangleq\mathcal{A}^{(t)}{}'\mathcal{A}^{(t)}(\boldsymbol{z})/(2n)\ .
\end{align*}
Notice when $n\rightarrow\infty$,
\begin{align*}
\mathcal{P}^{(t,0)}(\hat{\boldsymbol{y}}^{(t)}-\boldsymbol{y}^{(t)})\approx & \mathrm{tr}(M^{(t)}-M^{*})\\
\mathcal{P}^{(t,1)}(\hat{\boldsymbol{y}}^{(t)}-\boldsymbol{y}^{(t)})\approx & D(M^{(t)}-M^{*})\boldsymbol{\kappa}+\boldsymbol{w}^{(t)}-\boldsymbol{w}^{*}\\
\mathcal{P}^{(t,2)}(\hat{\boldsymbol{y}}^{(t)}-\boldsymbol{y}^{(t)})\approx & D(M^{(t)}-M^{*})(\boldsymbol{\phi}-1)\\
 & +D(\boldsymbol{\kappa})(\boldsymbol{w}^{(t)}-\boldsymbol{w}^{*})\ .
\end{align*}
 This inspires us to find a linear combination of $\mathcal{P}^{(t,\cdot)}$
to eliminate $F^{(t)}$. The solution for this linear combination
equation is
\begin{align}
\mathcal{M}^{(t)}(\hat{\boldsymbol{y}}^{(t)}-\boldsymbol{y}^{(t)})\triangleq & \mathcal{H}^{(t)}(\hat{\boldsymbol{y}}^{(t)}-\boldsymbol{y}^{(t)})\label{eq:mathcal-M-t}\\
 & -\frac{1}{2}\mathcal{D}\left(G_{1}\circ\mathcal{P}^{(t,1)}(\hat{\boldsymbol{y}}^{(t)}-\boldsymbol{y}^{(t)})\right)\nonumber \\
 & -\frac{1}{2}\mathcal{D}\left(G_{2}\circ\mathcal{P}^{(t,2)}(\hat{\boldsymbol{y}}^{(t)}-\boldsymbol{y}^{(t)})\right)\nonumber
\end{align}
 where
\begin{align}
G_{j,:}{}^{\top}= & \left[\begin{array}{cc}
1 & \boldsymbol{\kappa}_{j}\\
\boldsymbol{\kappa}_{j} & \boldsymbol{\phi}_{j}-1
\end{array}\right]^{-1}\left[\begin{array}{c}
\boldsymbol{\kappa}_{j}\\
\boldsymbol{\phi}_{j}-3
\end{array}\right]\label{eq:G}\\
H_{j,:}{}^{\top}= & \left[\begin{array}{cc}
1 & \boldsymbol{\kappa}_{j}\\
\boldsymbol{\kappa}_{j} & \boldsymbol{\phi}_{j}-1
\end{array}\right]^{-1}\left[\begin{array}{c}
1\\
0
\end{array}\right]\ .\label{eq:H}
\end{align}
 Similarly to eliminate the high order moments in the gradient of
$\boldsymbol{w}^{(t)}$ , we construct
\begin{align}
\mathcal{W}^{(t)}(\hat{\boldsymbol{y}}^{(t)}-\boldsymbol{y}^{(t)}) & \triangleq H_{1}\circ\mathcal{P}^{(t,1)}(\hat{\boldsymbol{y}}^{(t)}-\boldsymbol{y}^{(t)})\label{eq:mathcal-W-t}\\
 & +H_{2}\circ\mathcal{P}^{(t,2)}(\hat{\boldsymbol{y}}^{(t)}-\boldsymbol{y}^{(t)})\ .\nonumber
\end{align}

The overall construction is given in Algorithm \ref{alg:moment-estimation-sequence-method}.

We briefly prove that the construction in Algorithm \ref{alg:moment-estimation-sequence-method}
will eliminate the high order moments in $F^{(t)}$ by which a global
linear convergence rate is immediately followed. Please check appendix
for details. We will omit the superscript $(t)$ in $X^{(t)}$ and
$\mathcal{P}^{(t,\cdot)}$ when not raising confusion.

First we show that $\frac{1}{n}\mathcal{A}'\mathcal{A}$ is conditionally
independent restrict isometric after shifting its expectation (Shift
CI-RIP). The proof can be found in Appendix \ref{sec:Proof-of-Sub-gaussian-shifted-CI-RIP}.
\begin{thm}[Shift CI-RIP]
\label{thm:subgaussian-shifted-CI-RIP}  Suppose $d\geq(2+\|\boldsymbol{\phi}^{*}-3\|_{\infty})^{2}$.
Fixed a rank-$k$ matrix $M$, with probability at least $1-\eta$,
\begin{align*}
\frac{1}{n}\mathcal{A}'\mathcal{A}(M)= & 2M+\mathrm{tr}(M)I+\mathcal{D}(\boldsymbol{\phi}^{*}-3)\mathcal{D}(M)\\
 & +O(\delta\|M\|_{2})
\end{align*}
 provided $n\geq c_{\eta}(\sqrt{k}+|\mathrm{tr}(M)|)^{2}d/\delta^{2}$.
\end{thm}

Theorem \ref{thm:subgaussian-shifted-CI-RIP} is the main theorem
in our analysis. The key ingredient of our proof is to apply the matrix
Bernstein's inequality with an improved version of sub-gaussian Hanson-Wright
inequality proved by \citet{rudelson_hanson-wright_2013}. Please
check Appendix \ref{sec:Proof-of-Sub-gaussian-shifted-CI-RIP} for
more details.

Based on the shifted CI-RIP condition of operator $\mathcal{A}$,
we prove the following perturbation bounds.
\begin{lem}
\label{lem:concentration-of-mathcal-Pt} For $n\geq C_{\eta}(\sqrt{k}+|\mathrm{tr}(M)|)^{2}d/\delta^{2}$
, with probability at least $1-\eta$ ,
\begin{align*}
 & \frac{1}{n}\mathcal{A}'(X{}^{\top}\boldsymbol{w})=\mathcal{D}(\boldsymbol{\kappa}^{*})\boldsymbol{w}+O(\delta\|\boldsymbol{w}\|_{2})\\
 & \mathcal{P}^{(0)}(\boldsymbol{y})\triangleq\frac{1}{n}\boldsymbol{1}{}^{\top}\boldsymbol{y}=\mathrm{tr}(M)+O[\delta(\|\boldsymbol{w}\|_{2}+\|M\|_{2})]\\
 & \mathcal{P}^{(1)}(\boldsymbol{y})\triangleq\frac{1}{n}X\boldsymbol{y}=\mathcal{D}(M)\boldsymbol{\kappa}^{*}+\boldsymbol{w}+O[\delta(\|\boldsymbol{w}\|_{2}+\|M\|_{2})]\\
 & \mathcal{P}^{(2)}(\boldsymbol{y})\triangleq\frac{1}{n}X^{2}\boldsymbol{y}-\mathcal{P}^{(0)}(\boldsymbol{y})=\mathcal{D}(M)(\boldsymbol{\phi}^{*}-1)\\
 & \qquad+\mathcal{D}(\boldsymbol{\kappa}^{*})\boldsymbol{w}+O[\delta(\|\boldsymbol{w}\|_{2}+\|M\|_{2})]\ .
\end{align*}
\end{lem}

Lemma \ref{lem:concentration-of-mathcal-Pt} shows that $\mathcal{A}'X{}^{\top}$
and $\mathcal{P}^{(t,\cdot)}$ are all concentrated around their expectations
with no more than $O(C_{\eta}k^{2}d)$ samples. To finish our construction,
we need to bound the deviation of $G$ and $H$ from their expectation
$G^{*}$ and $H^{*}$ . This is done in the following lemma.
\begin{lem}
\label{lem:error-bound-G_j-G_j_start} Suppose the distribution of
$\boldsymbol{x}$ is $\tau$-MIP with $\tau>0$. Then in Algorithm
\ref{alg:moment-estimation-sequence-method},
\begin{align*}
\|G-G^{*}\|_{\infty}\leq & \delta,\ \|H-H^{*}\|_{\infty}\leq\delta\ ,
\end{align*}
 provided
\[
n\geq C_{\eta}(1+\tau^{-1}\sqrt{\|\boldsymbol{\kappa}^{*}\|_{\infty}^{2}+\|\boldsymbol{\phi}^{*}-3\|_{\infty}^{2}})/(\tau\delta^{2})\ .
\]
\end{lem}

Lemma \ref{lem:error-bound-G_j-G_j_start} shows that $G\approx G^{*}$
as long as $n\geq O(1/\tau^{2})$. The matrix inversion in the definition
of $G$ requires that the $\tau$-MIP condition must be satisfied
with $\tau>0$.

We are now ready to show that $\mathcal{M}^{(t)}$ and $\mathcal{W}^{(t)}$
are almost isometric.
\begin{lem}
\label{lem:concentration-of-mathcal-M-and-W} Under the same settings
of Theorem \ref{thm:global-convergence-rate-of-Moment-Estimation-Sequence},
with  probability at least $1-\eta$ ,
\begin{align*}
 & \mathcal{M}^{(t)}(\hat{\boldsymbol{y}}^{(t)}-\boldsymbol{y}^{(t)})=M^{(t-1)}-M^{*}+O(\delta\epsilon_{t-1})\\
 & \mathcal{W}^{(t)}(\hat{\boldsymbol{y}}^{(t)}-\boldsymbol{y}^{(t)})=\boldsymbol{w}^{(t-1)}-\boldsymbol{w}^{*}+O(\delta\epsilon_{t-1})
\end{align*}
provided
\[
n\geq C_{\eta}(p+1)^{2}/\delta^{2}\max\{p/\tau^{2},(\sqrt{k}+|\mathrm{tr}(M)|/\|M\|_{2})^{2}d\}
\]
 where $p\triangleq\max\{1,\|\boldsymbol{\kappa}^{*}\|_{\infty},\|\boldsymbol{\phi}^{*}-3\|_{\infty},\|\boldsymbol{\phi}^{*}-1\|_{\infty}\}$
.
\end{lem}

Lemma \ref{lem:concentration-of-mathcal-M-and-W} shows that $\mathcal{M}^{(t)}$
and $\mathcal{W}^{(t)}$ are almost isometric when the number of samples
is larger than $\mathcal{O}(k^{2}d)$ and $\mathcal{O}(1/\tau^{2})$.
The proof of Lemma \ref{lem:concentration-of-mathcal-M-and-W} consists
of two steps. First we replace each operator or matrix with its expectation
plus a small perturbation given in Lemma \ref{lem:concentration-of-mathcal-Pt}
and Lemma \ref{lem:error-bound-G_j-G_j_start}. Then Lemma \ref{lem:concentration-of-mathcal-M-and-W}
follows after simplification. Theorem \ref{thm:global-convergence-rate-of-Moment-Estimation-Sequence}
is obtained by combining Lemma \ref{lem:concentration-of-mathcal-M-and-W}
with alternating gradient descent analysis. Please check Appendix
\ref{sec:proof-of-theorem-global-convergence} for the complete proof.

\subsection{Improved Factorization Machine}

Theorem \ref{thm:global-convergence-rate-of-Moment-Estimation-Sequence}
shows that learning the generalized FM is hard on non-gaussian distribution.
Especially, when the data distribution has a very small $\tau$-MIP
constant, the sampling complexity to recover $M^{*}$ in Eq. (\ref{eq:gFM-model})
will be as large as $\mathcal{O}(1/\tau^{2})$. The recovery is even
impossible on $\tau=0$ distributions such as the Bernoulli distribution.
Clearly, a well-defined learnable model should not depend on the $\tau$-MIP
condition.

Indeed the bound given by Theorem \ref{thm:global-convergence-rate-of-Moment-Estimation-Sequence}
is quite sharp. It explains well why we cannot recover $M^{*}$ on
the Bernoulli distribution. Therefore it is unlikely to remove the
$\tau$ dependency by designing a better elimination sequence in Algorithm
\ref{alg:moment-estimation-sequence-method}. After examining the
proof of Theorem \ref{alg:moment-estimation-sequence-method} carefully,
we find that the only reason our bound contains $\tau$ is that the
diagonal elements of $M^{*}$ are allowed to be non-zero. If we constrain
$\mathcal{D}(M^{*})=0$ and $\mathcal{D}(M^{(t)})=0$, the $F^{(t)}$
in the expected gradient $\mathbb{E}\nabla_{V}\mathcal{L}(\boldsymbol{w}^{(t)},U^{(t)},V^{(t)})$
will be zero and then we do not need to eliminate it during the alternating
iteration. This greatly simplifies our convergence analysis as we
only need Theorem \ref{thm:subgaussian-shifted-CI-RIP} which now
becomes
\begin{align}
\frac{1}{n}\mathcal{A}'\mathcal{A}(M)= & 2M+O(\delta\|M\|_{2})\ .\label{eq:AA-RIP-when-Diag-M=00003D0}
\end{align}
Eq. (\ref{thm:subgaussian-shifted-CI-RIP}) already shows that $\frac{1}{n}\mathcal{A}'\mathcal{A}$
is almost isometric that immediately implies the linear convergence
rate of alternating gradient descent. As a direct corollary of Theorem
\ref{thm:subgaussian-shifted-CI-RIP}, the sampling complexity could
be improved to $\mathcal{O}(c_{\eta}kd)$ which is optimal up to some
logarithmic constants $c_{\eta}$. Inspired by these observations,
we propose to learn the following FM model
\begin{align}
 & y=\boldsymbol{x}{}^{\top}\boldsymbol{w}^{*}+\boldsymbol{x}{}^{\top}M^{*}\boldsymbol{x}\label{eq:iFM-model}\\
\mathrm{s.t.}\  & M^{*}=U^{*}V^{*}{}^{\top}-\mathcal{D}(U^{*}V^{*}{}^{\top})\ .\nonumber
\end{align}
 We called the above model the Improved Factorization Machine (iFM).
The iFM model is a trade-off between the conventional FM model and
the generalized FM model. It decouples the PSD constraint with $U\not=V$
in the generalized FM model but keeps the diagonal-zero constraint
as the conventional FM model. Unlike the conventional FM model, the
iFM model is proposed in a theoretical-driven way. The decoupling
of $\{U,V\}$ makes the iFM easy to optimize while the diagonal-zero
constraint makes it learnable with the optimal $\mathcal{O}(kd)$
sampling complexity. In the next section, we will verify the above
discussion with numerical experiments.

\section{Experiments}

We first use synthetic data in subsection \ref{subsec:Synthetic-Data}
to show the modeling power of iFM and the PSD bias of the conventional
FM. In subsection \ref{subsec:vTEC-Estimation} we apply iFM in a
real-word problem, the vTEC estimation task, to demonstrate its superiority
over baseline methods.

\subsection{Synthetic Data \label{subsec:Synthetic-Data}}

\begin{figure*}
\begin{minipage}[b][][b]{0.32 \linewidth}
\includegraphics[width=\linewidth]{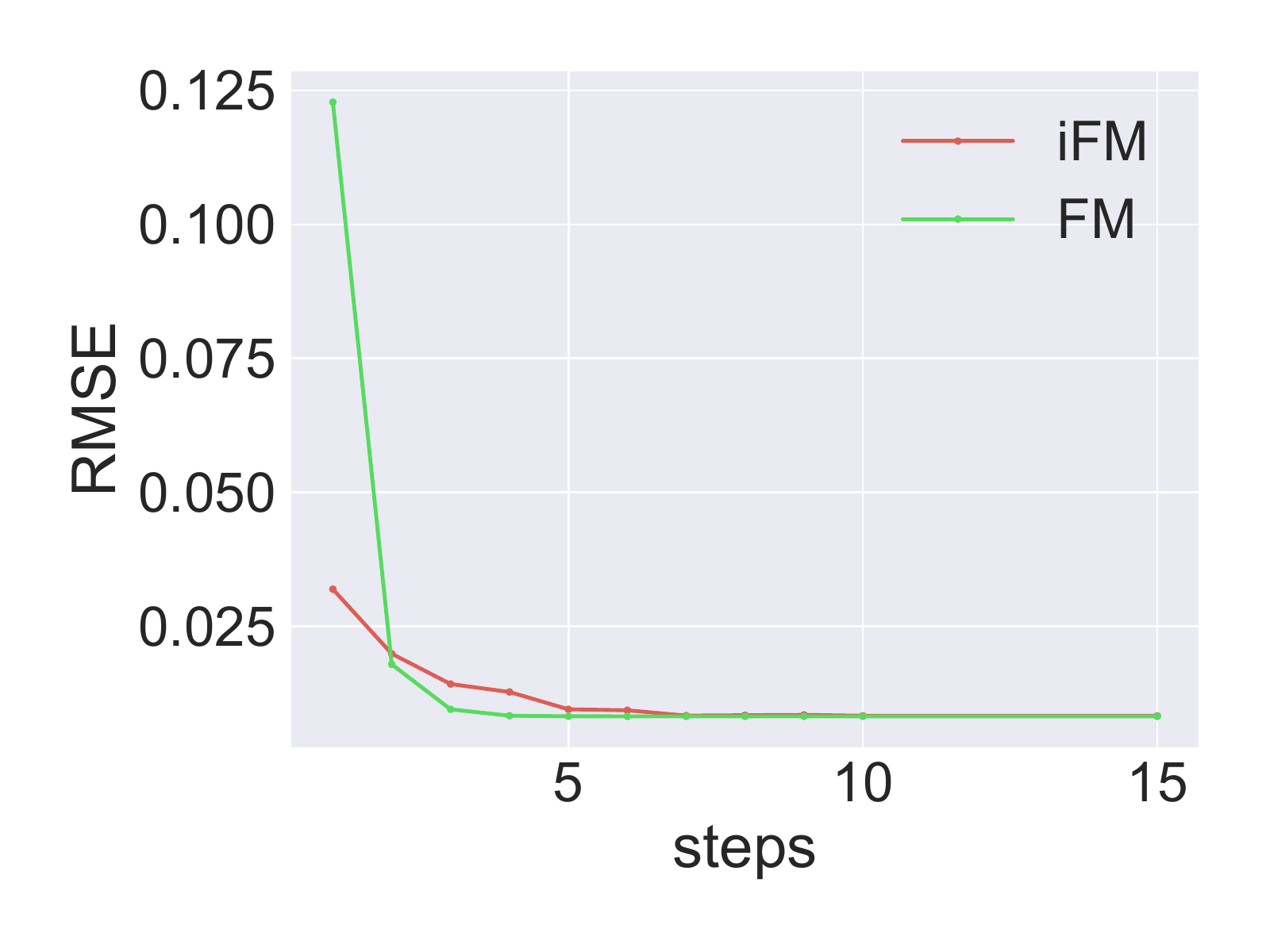}\\
\centering{(a) $M^*=U^* U^{*\top}-\mathrm{diag}( U^* U^{*\top} )$}
\end{minipage} \hfil
\begin{minipage}[b][][b]{0.32 \linewidth}
\includegraphics[width=\linewidth]{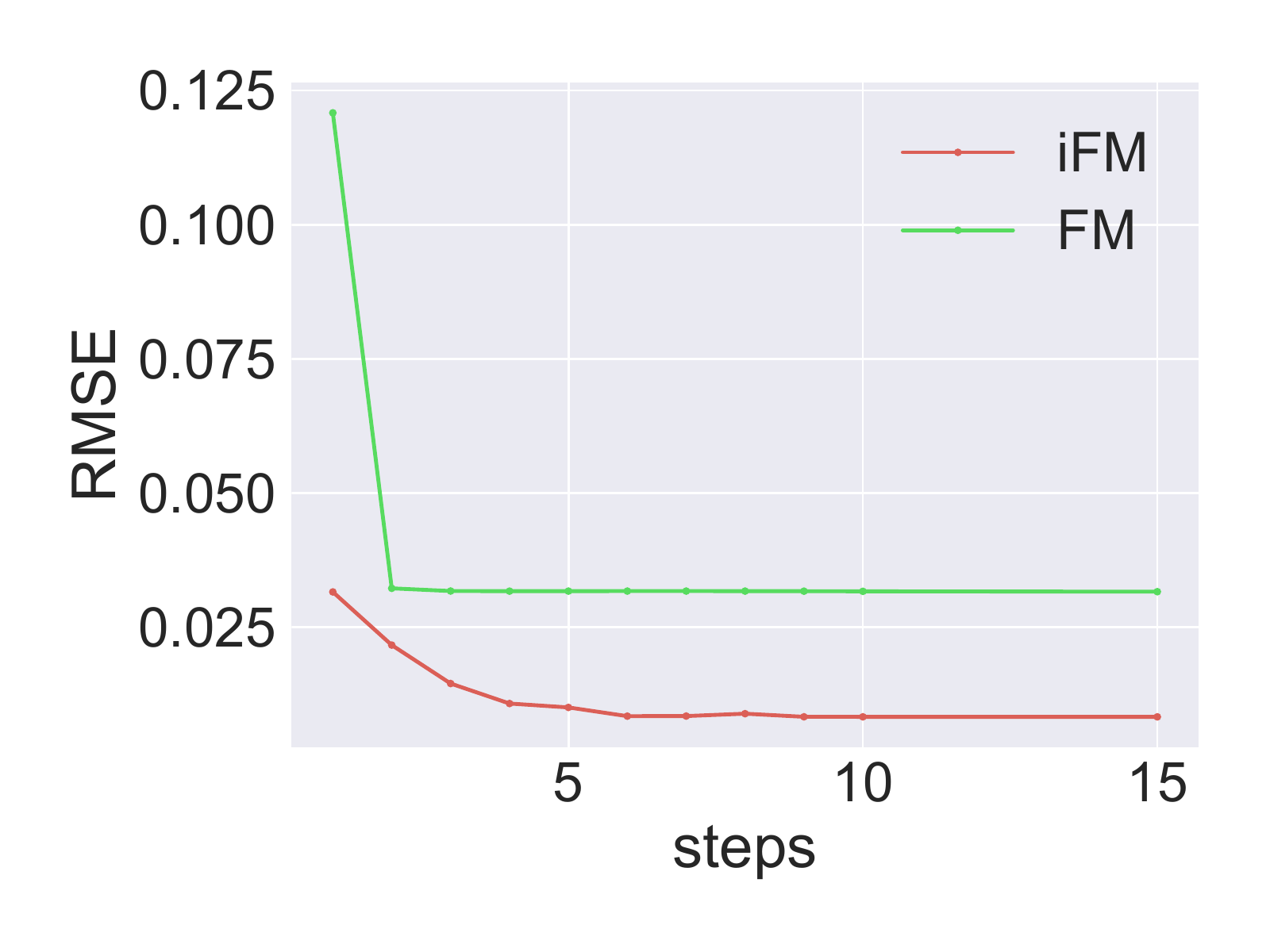}\\
\centering{(b) inverse the label $y$ in (a)}
\end{minipage} \hfil
\begin{minipage}[b][][b]{0.32 \linewidth}
\includegraphics[width=\linewidth]{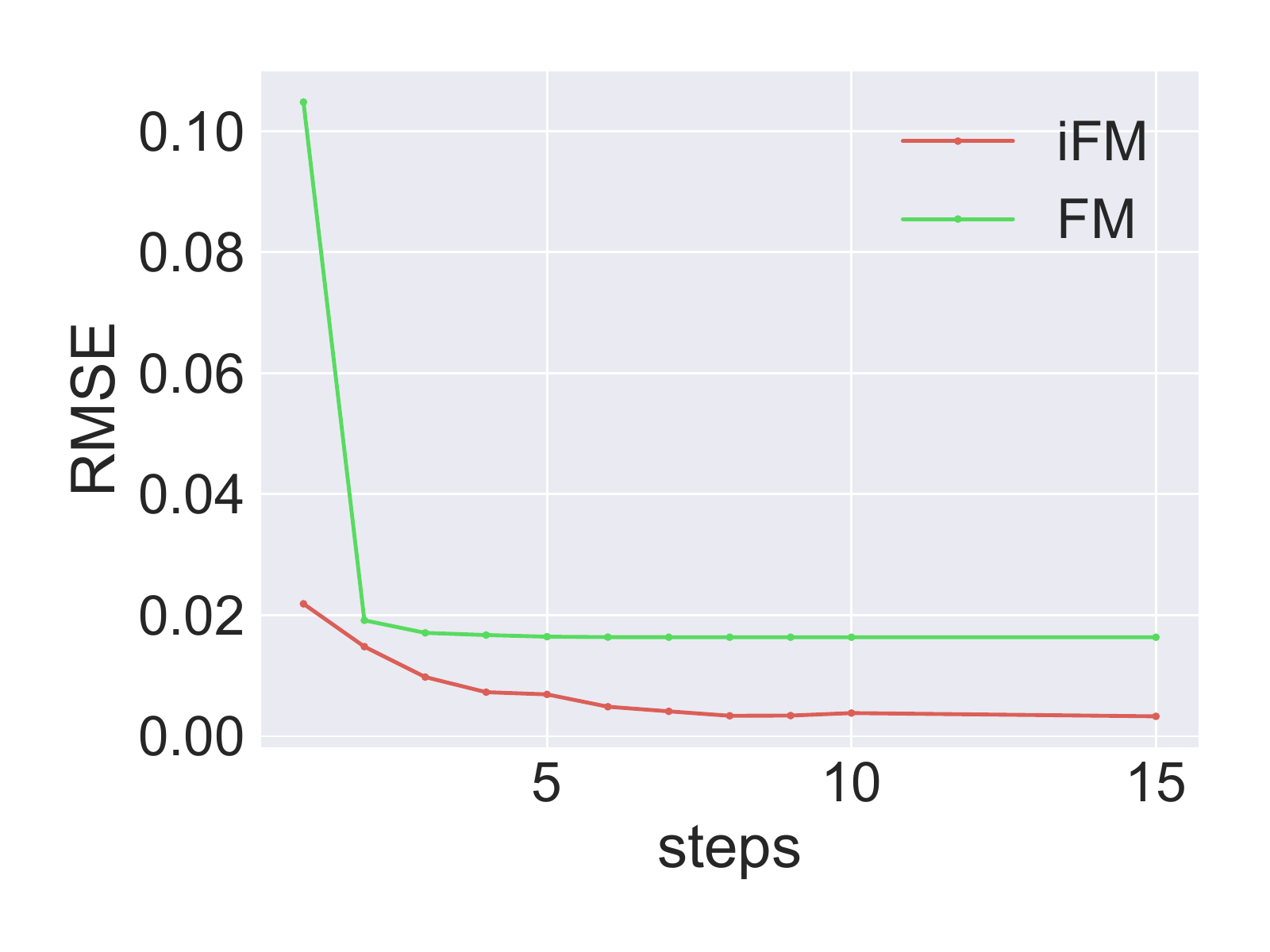}\\
\centering{(c) $M^*=U^* V^{*\top}-\mathrm{diag}( U^* V^{*\top} )$}
\end{minipage}

\caption{RMSE Curve of iFM v.s. FM}
\label{fig:test-rmse-curve}

\end{figure*}

In this subsection, we construct numerical examples to support our
theoretical results. To this end, we choose $d=100$, $k=5$. $\{\boldsymbol{w}^{*},U^{*},V^{*},\boldsymbol{x}\}$
are all sampled from the Gaussian distribution with variance $1/d$.
We randomly sample $30kd$ instances as training set and $10000$
instances as testing set. In Figure \ref{fig:test-rmse-curve}, we
report the convergence curve of iFM and FM on the testing set averaged
over 10 trials. The x-axis is the iteration step and the y-axis is
the Root Mean Square Error (RMSE) of $y$. In Figure (a), we generate
label $y$ following the conventional FM assumption. Both iFM and
FM converge well. In Figure (b), we flip the sign of the label in
(a). While iFM still converges well, FM cannot model the sign flip.
This example shows why we should avoid to use the conventional FM
both in theory and in practice: even a simple flipping operation can
make the model under-fit the data. In Figure (c), we generate $y$
from $M^{*}$ with both positive eigenvalues and negative eigenvalues.
Again the conventional FM cannot fit this data as the distribution
of $y$ is now symmetric in both direction.

\subsection{vTEC Estimation\label{subsec:vTEC-Estimation}}

\begin{table*}
\caption{RMSE \& RTK of iFM v.s. Baseline Methods}

\begin{centering}
\begin{tabular}{ccccccc}
\hline
 & Ridge & LASSO & ElasticNet & Kernel & FM & iFM\tabularnewline
\hline
TestDay1 & \bf{0.02161}  & 0.02260  & 0.02222  & 0.02570  & 0.02178  & 0.02164 \tabularnewline
TestDay2 & 0.01430  & 0.01461  & 0.01454  & 0.01683  & 0.01439  & \bf{0.01404} \tabularnewline
TestDay3 & 0.01508  & 0.01524  & 0.01496  & 0.01875  & \bf{0.01484 } & \bf{0.01484 }\tabularnewline
TestDay4 & 0.01449  & 0.01487  & 0.01460  & 0.01564  & 0.01432  & \bf{0.01423 }\tabularnewline
TestDay5 & 0.01610  & 0.01579  & 0.01612  & 0.01744  & 0.01606  & \bf{0.01567 }\tabularnewline
TestDay6 & 0.01487  & 0.01491  & 0.01483  & 0.01684  & 0.01470  & \bf{0.01459 }\tabularnewline
TestDay7 & 0.01828  & 0.01849  & 0.01827  & 0.02209  & 0.01830  & \bf{0.01805 }\tabularnewline
TestDay8 & 0.01461  & 0.01552  & 0.01519  & 0.01629  & 0.01466  & \bf{0.01453 }\tabularnewline
TestDay9 & 0.01657  & 0.01646  & 0.01639  & 0.02096  & 0.01646  & \bf{0.01636 }\tabularnewline
\hline
Average & 0.01621  & 0.01650  & 0.01635  & 0.01895  & 0.01617  & \bf{0.01599 }\tabularnewline
\hline
RTK  & 62.67\% & 62.18\% & 63.07\% & 52.09\% & 63.10\% & \bf{64.44\%}\tabularnewline
\hline
\end{tabular}
\par\end{centering}
\label{tab:RMSE}
\end{table*}

In this subsection, we demonstrate the superiority of iFM in a real-world
application, the \emph{vertical Total Electron Content} (vTEC) estimation.
The vTEC is an important descriptive parameter of the ionosphere of
the Earth. It integrates the total number of electrons integrated
when traveling from space to earth with a perpendicular path. One
important application is the real-time high precision GPS signal calibration.
The accuracy of the GPS system heavily depends on the pseudo-range
measurement between the satellite and the receiver. The major error
in the pseudo-range measurement is caused by the vTEC which is dynamic.

In order to estimate the vTEC, we build a triangle mesh grid system
in an anonymous region. Each node in the grid is a ground stations
equipped with dual-frequency high precision GPS receiver. The distance
between two nodes is around 75 kilometers. The station sends out resolved
GPS data every second. Formally, our system solves an online regression
problem. Our system receives $N_{t}$ data points at every time step
$t$ measured in seconds. Each data point $\boldsymbol{x}^{(t,i)}\in\mathbb{R}^{4}$
presents an ionospheric pierce point of a satellite-receiver pair.
The first two dimensions are the latitude $\alpha^{(t,i)}$ and the
longitude $\beta^{(t,i)}$ of the pierce point. The third dimension
and the fourth dimension are the zenith angle $\theta^{(t,i)}$ and
the azimuth angle $\gamma^{(t,i)}$ respectively. We will omit the
superscript $t$ below as we always stay within the same time window
$t$. In order to build a localized prediction model, we encode $\{\alpha^{(i)},\beta^{(i)}\}$
into a high dimensional vector. Suppose we have $m$ satellites in
total. First we collect data points for 60 seconds. Then we cluster
the collected $\{\alpha^{(i)},\beta^{(i)}\}$ into $m$ clusters via
K-means algorithm. Denote the cluster center of K-means as $\{\boldsymbol{c}^{(1)},\boldsymbol{c}^{(2)},\cdots,\boldsymbol{c}^{(m)}\}$
and the $i$-th data point belongs to the $g_{i}$-th cluster. The
first two dimensions of the $i$-th data point $\{\alpha^{(i)},\beta^{(i)}\}$
are then encoded as $\boldsymbol{v}^{(i)}\in\mathbb{R}^{2m}$ where
\begin{align*}
\boldsymbol{v}_{j}^{(i)}= & \begin{cases}
\{\alpha^{(i)},\beta^{(i)}\}-\boldsymbol{c}^{(g_{i})} & j=g_{i}\\
0 & \mathrm{otherwise}
\end{cases}
\end{align*}
 Finally, each data point $\boldsymbol{x}^{(i)}$ is encoded into
an $\mathbb{R}^{2m+8}$ vector:
\begin{align*}
\mathrm{Enc}(\boldsymbol{x}^{(i)})\triangleq & [\boldsymbol{v}^{(i)},\sin(\theta^{(i)}),\cos(\theta^{(i)}),\theta^{(i)},(\theta^{(i)})^{2},\\
 & \sin(\gamma^{(i)}),\cos(\gamma^{(i)}),\gamma^{(i)},(\gamma^{(i)})^{2}]\ .
\end{align*}

\paragraph{Evaluation}

It is important to note that our problem does not fit the conventional
machine learning framework. We only have validation set for model
selection and evaluation set to evaluate the model performance. We
introduce the training station and the testing station that correspond
to the ``training set'' and ``testing set'' in the conventional
machine learning framework. However please be advised that they are
not exactly the same concepts.

To evaluate the performance of our models, we randomly select one
ground station as the testing station. Around the testing station,
we choose $16$ ground stations as training stations to learn the
online prediction model. Suppose the online prediction model is $\mathcal{F}$
which maps $\mathrm{Enc}(\boldsymbol{x}^{(i)})$ to the corresponding
vTEC value
\[
\mathcal{F}:\mathrm{Enc}(\boldsymbol{x}^{(i)})\rightarrow\mathrm{vTEC}(\boldsymbol{x}^{(i)})\in\mathbb{R}\ .
\]
 In our system, we are only given the double difference of the $\mathrm{vTEC}$
values due to the signal resolving process. Suppose two satellites
$a,b$ and two ground stations $c,d$ are connected in the data-link
graph, $\boldsymbol{x}^{(a,c)}$ denotes the ionospheric pierce point
between $a$ and $c$. The observed double difference $y^{(a,b,c,d)}$
is given by
\begin{align*}
y^{(a,b,c,d)}\triangleq & \mathrm{vTEC}(\boldsymbol{x}^{(a,c)})-\mathrm{vTEC}(\boldsymbol{x}^{(a,d)})-\mathrm{vTEC}(\boldsymbol{x}^{(b,c)})\\
 & +\mathrm{vTEC}(\boldsymbol{x}^{(b,d)})\ .
\end{align*}
 Since $\mathrm{vTEC}(\cdot)$ is an unknown function, we need to
approximate it by $\mathcal{F}$. Once $\mathcal{F}$ is learned from
training stations, we can apply it to predict the double difference
$y^{(a,b,c,d)}$ where either $c$ or $d$ is the testing station.

Once we get the vTEC estimation, we use it to calibrate the GPS signal
and finally compute the geometric coordinate of the user. The RTK
ratio measures the quality of the positioning service. It is a real
number presenting the probability of successful positioning with accuracy
at least one centimeter. The RTK ratio is computed from a commercial
system that is much slower than the computation of RMSE.

\paragraph{Dataset and Results}

We select a ground station at the region center as testing station.
Around the testing station 16 stations are selected as training stations.
We collect 5 consecutive days' data as validation set for parameter
tuning. The following 9 days' data are used as evaluation set. We
update the prediction model per 60 seconds. The learned model is then
used to predict the double differences relating to the testing station.
We compare the predicted double differences to the true values detected
by the testing station. The number of valid satellites in our experiment
is around 10 to 20.

In Table \ref{tab:RMSE}, we report the root-mean-square error (RMSE)
over 9 days period. The dates are denoted as TestDay1 to TestDay9
for anonymity. Five baseline methods are evaluated: Ridge Regression,
LASSO, ElasticNet, Kernel Ridge Regression (Kernel) with RBF kernel
and the conventional Factorization Machine (FM). More computational
expensive models such as deep neural network are not feasible for
our online system. For Ridge, LASSO, Kernel and ElasticNet, their
parameters are tuned from $1\times10^{-6}$ to $1\times10^{6}$. The
regularizer parameters of FM and iFM are tuned from $1\times10^{-6}$
to $1\times10^{6}$. The rank of $M$ is tuned in set $\{1,2,\cdots,10\}$.
We use Scikit-learn \cite{scikit-learn} and fastFM \cite{bayer_fastfm:_2016}
to implement the baseline methods.

In Table \ref{tab:RMSE}, we observe that iFM is uniformly better
than the baseline methods. We average the root squared error over
$9\times24\times60=12960$ minutes in the last second row. The 95\%
confidence interval is within $1\times10^{-5}$ in our experiment.
In our experiment, the optimal rank of FM is 2 and the optimal rank
of iFM is 6. We note that FM is better than the first order linear
models since it captures the second order information. This indicates
that the second order information is indeed helpful.

In the last row of Table \ref{tab:RMSE}, we report the RTK ratio
averaged over the 9 days. We find that the RTK ratio will improve
a lot even with small improvement of vTEC estimation. This is because
the error of vTEC estimation will be broadcasted and magnified in
the RTK computation pipeline. The RTK ratio of iFM is about 1.77\%
better than that of Ridge regression and is more than 12\% better
than Kernel regression. Comparing to FM, it is 1.34\% better. We conclude
that iFM achieves overall better performance and the improvement is
statistically significant.

\section{Conclusion}

We study the learning guarantees of the FM solved by alternating gradient
descent on sub-gaussian distributions. We find that the conventional
modeling of the factorization machine might be sub-optimal in capturing
negative second order patterns. We prove that the constraints in the
conventional FM can be removed resulting a generalized FM model learnable
by $\max\{\mathcal{O}(k^{2}d\},\mathcal{O}(1/\tau^{2})\}$ samples.
The sampling complexity can be improved to the optimal $\mathcal{O}(kd)$
with diagonal-zero constraint. Our theoretical analysis shows that
the optimal modeling of high order linear model does not always agree
with the heuristic intuition. We hope this work could inspire future
researches of non-convex high order machines with solid theoretical
foundation.

\section{ Acknowledgments}
This research is supported in part by NSF (III-1539991).
The high precision GPS dataset is provided by Qianxun Spatial Intelligence Inc. China. We appreciate Dr. Wotao Yin from University of California Los Angeles and anonymous reviewers for their insightful comments.

\bibliographystyle{plain}
\bibliography{refs}

\newpage

\appendix
\onecolumn

\section{Preliminary}

The $\psi_{2}$-Orlicz norm of a random sub-gaussian variable $z$
is defined by
\[
\|z\|_{\psi_{2}}\triangleq\inf\{t>0:\mathbb{E}\exp(z^{2}/t^{2})\leq c\}
\]
 where $c>0$ is a constant. For a random sub-gaussian vector $\boldsymbol{z}\in\mathbb{R}^{n}$,
its $\psi_{2}$-Orlicz norm is
\[
\|\boldsymbol{z}\|_{\psi_{2}}\triangleq\sup_{\boldsymbol{x}\in S^{n-1}}\|\left\langle \boldsymbol{z},\boldsymbol{x}\right\rangle \|_{\psi_{2}}
\]
 where $S^{n-1}$ is the unit sphere.

The following theorem gives the matrix Bernstein's inequality \cite{roman_vershynin_high-dimensional_2017}.
\begin{thm}[Matrix Bernstein's inequality]
 \label{thm:matrix-berstein} Let $X_{1},\cdots,X_{N}$ be independent,
mean zero $d\times n$ random matrices with $d\geq n$ and $\|X_{i}\|_{2}\leq B$.
Denote
\[
\sigma^{2}\triangleq\max\{\|\sum_{i=1}^{N}\mathbb{E}X_{i}X_{i}{}^{\top}\|_{2},\|\sum_{i=1}^{N}\mathbb{E}X_{i}{}^{\top}X_{i}\|_{2}\}\ .
\]
 Then for any $t\geq0$, we have
\[
\mathbb{P}(\|\sum_{i=1}^{N}X_{i}\|_{2}\geq t)\leq2d\exp\left[-c\min\left(\frac{t^{2}}{\sigma^{2}},\frac{t}{B}\right)\right]\ .
\]
 where $c$ is a universal constant. Equivalently, with probability
at least $1-\eta$,
\[
\|\sum_{i=1}^{N}X_{i}\|_{2}\leq c\max\left\{ B\log(2d/\eta),\sigma\sqrt{\log(2d/\eta)}\right\} \ .
\]
 When $\mathbb{E}X_{i}\not=\boldsymbol{0}$, replacing $X_{i}$ with
$X_{i}-\mathbb{E}X_{i}$ the inequality still holds true.
\end{thm}
The following Hanson-Wright inequality for sub-gaussian variables
is given in \cite{rudelson_hanson-wright_2013} .
\begin{thm}[Sub-gaussian Hanson-Wright inequality]
 \label{thm:hanson-wright-inequality} Let $\boldsymbol{x}=[x_{1},\cdots,x_{d}]\in\mathbb{R}^{d}$
be a random vector with independent, mean zero, sub-gaussian coordinates.
Then given a fixed $d\times d$ matrix $M$, for any $t\geq0$,
\[
\mathbb{P}\left\{ |\boldsymbol{x}{}^{\top}A\boldsymbol{x}-\mathbb{E}\boldsymbol{x}{}^{\top}A\boldsymbol{x}|\geq t\right\} \leq2\exp\left[-c\min\left(\frac{t^{2}}{B^{4}\|A\|_{F}^{2}},\frac{t}{B^{2}\|A\|_{2}}\right)\right]\ ,
\]
 where $B=\max_{i}\|x_{i}\|_{\psi_{2}}$ and $c$ is a universal positive
constant. Equivalently, with probability at least $1-\eta$,
\begin{align*}
|\boldsymbol{x}{}^{\top}A\boldsymbol{x}-\mathbb{E}\boldsymbol{x}{}^{\top}A\boldsymbol{x}|\leq & c\max\{B^{2}\|A\|_{2}\log(2/\eta),B^{2}\|A\|_{F}\sqrt{\log(2/\eta)}\}\ .
\end{align*}
\end{thm}

\paragraph{Truncation trick}

As Bernstein's inequality requires boundness of the random variable,
we use the truncation trick in order to apply it on unbounded random
matrices. First we condition on the tail distribution of random matrices
to bound the norm of a fixed random matrix. Then we take union bound
over all $n$ random matrices in the summation. The union bound will
result in an extra $O[\log(n)]$ penalty in the sampling complexity
which can be absorbed into $C_{\eta}$ or $c_{\eta}$ . Please check
\citep{tao_topics_2012} for more details.

\section{Proof of Theorem \ref{thm:subgaussian-shifted-CI-RIP} \label{sec:Proof-of-Sub-gaussian-shifted-CI-RIP}}

Define $p_{1}=2+\|\boldsymbol{\phi}^{*}-3\|_{\infty}$ . Recall that
\begin{align*}
\frac{1}{n}\mathcal{A}'\mathcal{A}(M)= & \frac{1}{n}\sum_{i=1}^{n}\boldsymbol{x}^{(i)}\boldsymbol{x}^{(i)}{}^{\top}M\boldsymbol{x}^{(i)}\boldsymbol{x}^{(i)}{}^{\top}\ .
\end{align*}
 Denote
\begin{align*}
 & Z_{i}\triangleq\boldsymbol{x}^{(i)}\boldsymbol{x}^{(i)}{}^{\top}M\boldsymbol{x}^{(i)}\boldsymbol{x}^{(i)}{}^{\top}\\
 & \mathbb{E}Z_{i}=2M+\mathrm{tr}(M)I+\mathcal{D}(\boldsymbol{\phi}^{*}-3)\mathcal{D}(M)\ .
\end{align*}
 In order to apply matrix Bernstein's inequality , we have
\begin{align*}
\|Z_{i}\|_{2}= & \|\boldsymbol{x}^{(i)}\boldsymbol{x}^{(i)}{}^{\top}M\boldsymbol{x}^{(i)}\boldsymbol{x}^{(i)}{}^{\top}\|_{2}\\
\leq & |\boldsymbol{x}^{(i)}{}^{\top}M\boldsymbol{x}^{(i)}|\|\boldsymbol{x}^{(i)}\boldsymbol{x}^{(i)}{}^{\top}\|_{2}\\
\leq & |\boldsymbol{x}^{(i)}{}^{\top}M\boldsymbol{x}^{(i)}|\|\boldsymbol{x}^{(i)}\|_{2}^{2}\\
\leq & c_{\eta}[\|M\|_{F}+|\mathrm{tr}(M)|]\|\boldsymbol{x}^{(i)}\|_{2}^{2}\\
\leq & c_{\eta}[\|M\|_{F}+|\mathrm{tr}(M)|]d\ .
\end{align*}

The 3rd inequality is because the Hanson-Wright inequality and the fact that $\mathbb{E} \boldsymbol{x}^{(i)}{}^{\top}M\boldsymbol{x}^{(i)} = \mathrm{tr}(M)$ (See Appendix C, Proof of Lemma 5).

 And
\begin{align*}
\|\mathbb{E}Z_{i}\|_{2}= & \|2M+\mathrm{tr}(M)I+\mathcal{D}(\boldsymbol{\phi}^{*}-3)\mathcal{D}(M)\|_{2}\\
\leq & 2\|M\|_{2}+|\mathrm{tr}(M)|+\|\boldsymbol{\phi}^{*}-3\|_{\infty}\|M\|_{2}\\
\leq & (2+\|\boldsymbol{\phi}^{*}-3\|_{\infty})\|M\|_{2}+|\mathrm{tr}(M)|\\
\leq & p_{1}\|M\|_{2}+|\mathrm{tr}(M)|\ .
\end{align*}
The last inequality is because the definition of $p_1$.

 And
\begin{align*}
\|\mathbb{E}Z_{i}Z_{i}{}^{\top}\|_{2}= & \|\mathbb{E}\boldsymbol{x}^{(i)}\boldsymbol{x}^{(i)}{}^{\top}M\boldsymbol{x}^{(i)}\boldsymbol{x}^{(i)}{}^{\top}\boldsymbol{x}^{(i)}\boldsymbol{x}^{(i)}{}^{\top}M\boldsymbol{x}^{(i)}\boldsymbol{x}^{(i)}{}^{\top}\|_{2}\\
\leq & c_{\eta}d\|\mathbb{E}\boldsymbol{x}^{(i)}\boldsymbol{x}^{(i)}{}^{\top}M\boldsymbol{x}^{(i)}\boldsymbol{x}^{(i)}{}^{\top}M\boldsymbol{x}^{(i)}\boldsymbol{x}^{(i)}{}^{\top}\|_{2}\\
\leq & c_{\eta}d\|\mathbb{E}\boldsymbol{x}^{(i)}\boldsymbol{x}^{(i)}{}^{\top}\|_{2}|\boldsymbol{x}^{(i)}{}^{\top}M\boldsymbol{x}^{(i)}|^{2}\\
\leq & c_{\eta}d\|\mathbb{E}\boldsymbol{x}^{(i)}\boldsymbol{x}^{(i)}{}^{\top}\|_{2}[\|M\|_{F}+|\mathrm{tr}(M)|]^{2}\\
\leq & c_{\eta}d[\|M\|_{F}+|\mathrm{tr}(M)|]^{2}\ .
\end{align*}
 And
\begin{align*}
\|(\mathbb{E}Z_{i})(\mathbb{E}Z_{i}){}^{\top}\|_{2}\leq & \|\mathbb{E}Z_{i}\|_{2}^{2}\\
\leq & [p_{1}\|M\|_{2}+|\mathrm{tr}(M)|]^{2}\ .
\end{align*}
 Therefore we get
\begin{align*}
\|Z_{i}-\mathbb{E}Z_{i}\|_{2}\leq & \|Z_{i}\|_{2}+\|\mathbb{E}Z_{i}\|_{2}\\
\leq & c_{\eta}[\|M\|_{F}+|\mathrm{tr}(M)|]d+p_{1}\|M\|_{2}+|\mathrm{tr}(M)|\ .
\end{align*}
 And
\begin{align*}
\mathrm{Var}1\triangleq & \|\mathbb{E} (Z_{i}-\mathbb{E}Z_{i})(Z_{i}-\mathbb{E}Z_{i}){}^{\top}\|_{2}\\
\leq & \|Z_{i}Z_{i}{}^{\top}\|_{2}+\|(\mathbb{E}Z_{i})(\mathbb{E}Z_{i}){}^{\top}\|_{2}\\
\leq & c_{\eta}d[\|M\|_{F}+|\mathrm{tr}(M)|]^{2}+[p_{1}\|M\|_{2}+|\mathrm{tr}(M)|]^{2}\ .
\end{align*}
 Suppose that
\begin{align*}
 & d[\|M\|_{F}+|\mathrm{tr}(M)|]^{2}\geq[p_{1}\|M\|_{2}+|\mathrm{tr}(M)|]^{2}\\
\Leftarrow & d[\|M\|_{2}+|\mathrm{tr}(M)|]^{2}\geq[p_{1}\|M\|_{2}+|\mathrm{tr}(M)|]^{2}\\
\Leftarrow & d[\|M\|_{2}+|\mathrm{tr}(M)|]^{2}\geq p_{1}^{2}[\|M\|_{2}+|\mathrm{tr}(M)|]^{2}\\
\Leftarrow & d\geq p_{1}^{2}\ .
\end{align*}
 And suppose that
\begin{align*}
 & [\|M\|_{F}+|\mathrm{tr}(M)|]d\geq p_{1}\|M\|_{2}+|\mathrm{tr}(M)|\\
\Leftarrow & d\geq p_{1}\\
\Leftarrow & d\geq p_{1}^{2}\ .
\end{align*}
The we get
\begin{align*}
\|Z_{i}-\mathbb{E}Z_{i}\|_{2}\leq & c_{\eta}[\|M\|_{F}+|\mathrm{tr}(M)|]d\\
\leq & c_{\eta}(\sqrt{k}+|\mathrm{tr}(M)|/\|M\|_{2})d\|M\|_{2}\\
\mathrm{Var}1\leq & c_{\eta}d[\|M\|_{F}+|\mathrm{tr}(M)|]^{2}\\
\leq & c_{\eta}(\sqrt{k}+|\mathrm{tr}(M)|/\|M\|_{2})^{2}d\|M\|_{2}^{2}\ .
\end{align*}
 Then according to matrix Bernstein's inequality,
\begin{align*}
\|\frac{1}{n}\sum_{i=1}^{n}Z_{i}-\mathbb{E}Z_{i}\|_{2}= & c_{\eta}\max\{\frac{1}{n}(\sqrt{k}+|\mathrm{tr}(M)|/\|M\|_{2})d\|M\|_{2},\frac{1}{\sqrt{n}}\sqrt{k}+|\mathrm{tr}(M)|/\|M\|_{2}\sqrt{d}\|M\|_{2}\}\\
\leq & c_{\eta}\frac{1}{\sqrt{n}}(\sqrt{k}+|\mathrm{tr}(M)|/\|M\|_{2})\sqrt{d}\|M\|_{2}\ .
\end{align*}
 provided
\begin{align*}
 & \frac{1}{n}kd\|M\|_{2}\leq\frac{1}{\sqrt{n}}(\sqrt{k}+|\mathrm{tr}(M)|/\|M\|_{2})\sqrt{d}\|M\|_{2}\\
\Leftarrow & n\geq d\ .
\end{align*}
 Choose $n\geq c_{\eta}(\sqrt{k}+|\mathrm{tr}(M)|/\|M\|_{2})^{2}d/\delta^{2}$,
we get
\begin{align*}
\|\frac{1}{n}\sum_{i=1}^{n}Z_{i}-\mathbb{E}Z_{i}\|_{2}\leq & \delta\|M\|_{2}\ .
\end{align*}

\section{Proof of Lemma \ref{lem:concentration-of-mathcal-Pt}}
\begin{proof}
To prove $\frac{1}{n}\mathcal{A}'(X{}^{\top}\boldsymbol{w})$,
\begin{align*}
\frac{1}{n}\mathcal{A}'(X{}^{\top}\boldsymbol{w})= & \frac{1}{n}\sum_{i=1}^{n}\boldsymbol{x}^{(i)}\boldsymbol{x}^{(i)}{}^{\top}\boldsymbol{w}\boldsymbol{x}^{(i)}{}^{\top}\ .
\end{align*}
 Similar to Theorem \ref{thm:subgaussian-shifted-CI-RIP}, just replacing
$\mathcal{A}(M)$ with $\boldsymbol{w}$, then with probability at
last $1-\eta$,
\begin{align*}
\|\frac{1}{n}\mathcal{A}'(X{}^{\top}\boldsymbol{w})-\mathcal{D}(\boldsymbol{\kappa}^{*})\boldsymbol{w}\|_{2}\leq & C_{\eta}\sqrt{d/n}\|\boldsymbol{w}\|_{2}\ .
\end{align*}
 Therefore let
\begin{align*}
 & n\geq C_{\eta}d/\delta^{2}\ .
\end{align*}
 We have
\begin{align*}
 & \|\frac{1}{n}\mathcal{A}'(X{}^{\top}\boldsymbol{w})-\mathcal{D}(\boldsymbol{\kappa}^{*})\boldsymbol{w}\|_{2}\leq\delta\|\boldsymbol{w}\|_{2}\ .
\end{align*}

To prove $\mathcal{P}^{(0)}(\boldsymbol{y})$,
\begin{align*}
\mathcal{P}^{(0)}(\boldsymbol{y})= & \frac{1}{n}\sum_{i=1}^{n}\boldsymbol{x}^{(i)}{}^{\top}\boldsymbol{w}+\frac{1}{n}\sum_{i=1}^{n}\boldsymbol{x}^{(i)}{}^{\top}M\boldsymbol{x}^{(i)}\ .
\end{align*}
Since $\boldsymbol{x}$ is coordinate sub-gaussian, any $i\in\{1,\cdots,d\}$,
with probability at least $1-\eta$,
\[
\|\boldsymbol{x}^{(i)}{}^{\top}\boldsymbol{w}\|_{2}\leq c\sqrt{d}\|\boldsymbol{w}\|_{2}\log(n/\eta)\ .
\]
 Then we have
\begin{align*}
\|\frac{1}{n}\sum_{i=1}^{n}\boldsymbol{x}^{(i)}{}^{\top}\boldsymbol{w}-0\|_{2}\leq & C\sqrt{d}\|\boldsymbol{w}\|_{2}\log(n/\eta)/\sqrt{n}\ .
\end{align*}
 Choose $n\geq c_{\eta}d$, we get
\begin{align*}
\|\frac{1}{n}\sum_{i=1}^{n}\boldsymbol{x}^{(i)}{}^{\top}\boldsymbol{w}\|_{2}\leq & \delta\|\boldsymbol{w}\|_{2}\ .
\end{align*}
 From Hanson-Wright inequality,
\begin{align*}
\|\frac{1}{n}\sum_{i=1}^{n}\boldsymbol{x}^{(i)}{}^{\top}M\boldsymbol{x}^{(i)}-\mathrm{tr}(M)\|_{2}\leq & C\|M\|_{F}\log(1/\eta)\\
\leq & C\|M\|_{2}\sqrt{k/n}\log(1/\eta)\ .
\end{align*}
 Therefore
\begin{align*}
\mathcal{P}^{(0)}(\boldsymbol{y})= & \mathrm{tr}(M)+O[(\sqrt{d}\|\boldsymbol{w}\|_{2}+\|M\|_{2}\sqrt{k})/\sqrt{n}\log(n/\eta)]\\
= & \mathrm{tr}(M)+O[C_{\eta}(\sqrt{d}\|\boldsymbol{w}\|_{2}+\|M\|_{2}\sqrt{k})/\sqrt{n}]\\
= & \mathrm{tr}(M)+O[C_{\eta}(\|\boldsymbol{w}\|_{2}+\|M\|_{2}\sqrt{k})\sqrt{d/n}]\ .
\end{align*}
 Let
\begin{align*}
 & n\geq C_{\eta}kd/\delta^{2}\ .
\end{align*}
We have
\begin{align*}
\mathcal{P}^{(0)}(\boldsymbol{y})= & \mathrm{tr}(M)+O[\delta(\|\boldsymbol{w}\|_{2}+\|M\|_{2})]\ .
\end{align*}

To prove $\mathcal{P}^{(1)}(\boldsymbol{y})$,
\begin{align*}
\mathcal{P}^{(1)}(\boldsymbol{y})= & \frac{1}{n}\sum_{i=1}^{n}\boldsymbol{x}^{(i)}\boldsymbol{x}^{(i)}{}^{\top}\boldsymbol{w}+\frac{1}{n}\sum_{i=1}^{n}\boldsymbol{x}^{(i)}\boldsymbol{x}^{(i)}{}^{\top}M\boldsymbol{x}^{(i)}\ .
\end{align*}
 From co-variance concentration inequality,
\begin{align*}
\|\frac{1}{n}\sum_{i=1}^{n}\boldsymbol{x}^{(i)}\boldsymbol{x}^{(i)}{}^{\top}\boldsymbol{w}-\boldsymbol{w}\|_{2}\leq & c\sqrt{d/n}\|\boldsymbol{w}\|_{2}\log(d/\eta)\\
\leq & C_{\eta}\sqrt{d/n}\|\boldsymbol{w}\|_{2}\ .
\end{align*}
 To bound the second term in $\mathcal{P}^{(1)}(\boldsymbol{y})$,
apply Hanson-Wright inequality again,
\begin{align*}
\|\boldsymbol{x}^{(i)}\boldsymbol{x}^{(i)}{}^{\top}M\boldsymbol{x}^{(i)}\|_{2}\leq & \|\boldsymbol{x}^{(i)}\|_{2}\|\boldsymbol{x}^{(i)}{}^{\top}M\boldsymbol{x}^{(i)}\|_{2}\\
\leq & c[\|M\|_{F}+\mathrm{tr}(M)]\sqrt{d}\log^{2}(nd/\eta)\\
\leq & C_{\eta}(\sqrt{k}+|\mathrm{tr}(M)|/\|M\|_{2})\|M\|_{2}\sqrt{d}\ .
\end{align*}
 By matrix Chernoff's inequality, choose $n\geq c_{\eta}(\sqrt{k}+|\mathrm{tr}(M)|/\|M\|_{2})^{2}d/\delta^{2}$,
\begin{align*}
\|\frac{1}{n}\sum_{i=1}^{n}\boldsymbol{x}^{(i)}\boldsymbol{x}^{(i)}{}^{\top}M\boldsymbol{x}^{(i)}-\mathcal{D}(M)\boldsymbol{\kappa}^{*}\|_{2}\leq & C_{\eta}(\sqrt{k}+|\mathrm{tr}(M)|/\|M\|_{2})\|M\|_{2}\sqrt{d/n}\\
\leq & \delta\|M\|_{2}\ .
\end{align*}
 Therefore we have
\begin{align*}
\mathcal{P}^{(1)}(\boldsymbol{y})=\boldsymbol{w}+\mathcal{D}(M)\boldsymbol{\kappa}^{*}+O & [\delta(\|\boldsymbol{w}\|_{2}+\|M\|_{2})]\ .
\end{align*}

To bound $\mathcal{P}^{(2)}(\boldsymbol{y})$ , first note that
\begin{align*}
\mathcal{P}^{(2)}(\boldsymbol{y})= & \frac{1}{n}\sum_{i=1}^{n}\boldsymbol{x}^{(i)2}\boldsymbol{x}^{(i)}{}^{\top}\boldsymbol{w}+\frac{1}{n}\sum_{i=1}^{n}\boldsymbol{x}^{(i)2}\boldsymbol{x}^{(i)}{}^{\top}M\boldsymbol{x}^{(i)}-P^{(0)}(\boldsymbol{y})\\
= & \frac{1}{n}\sum_{i=1}^{n}\mathcal{D}(\boldsymbol{x}^{(i)}\boldsymbol{x}^{(i)}{}^{\top}\boldsymbol{w}\boldsymbol{x}^{(i)})+\frac{1}{n}\sum_{i=1}^{n}\mathcal{D}(\boldsymbol{x}^{(i)}\boldsymbol{x}^{(i)}{}^{\top}M\boldsymbol{x}^{(i)}\boldsymbol{x}^{(i)}{}^{\top})-P^{(0)}(\boldsymbol{y})\ .
\end{align*}
 Then similarly,
\begin{align*}
\|\frac{1}{n}\sum_{i=1}^{n}\boldsymbol{x}^{(i)2}\boldsymbol{x}^{(i)}{}^{\top}\boldsymbol{w}-\mathcal{D}(\boldsymbol{\kappa}^{*})\boldsymbol{w}\|_{2}\leq & C_{\eta}\sqrt{d/n}\|\boldsymbol{w}\|_{2}
\end{align*}
\begin{align*}
\|\frac{1}{n}\sum_{i=1}^{n}\boldsymbol{x}^{(i)2}\boldsymbol{x}^{(i)}{}^{\top}M\boldsymbol{x}^{(i)}-\mathrm{tr}(M)-\mathcal{D}(M)(\boldsymbol{\phi}^{*}-1)\|_{2}\leq & C_{\eta}(\sqrt{k}+|\mathrm{tr}(M)|/\|M\|_{2})\|M\|_{2}\sqrt{d/n}\ .
\end{align*}
The last inequality is because Theorem \ref{thm:subgaussian-shifted-CI-RIP}.
Combine all together, choose $n\geq c_{\eta}(\sqrt{k}+|\mathrm{tr}(M)|/\|M\|_{2})^{2}d$,
\begin{align*}
\mathcal{P}^{(2)}(\boldsymbol{y})= & \mathcal{D}(\boldsymbol{\kappa}^{*})\boldsymbol{w}+\mathcal{D}(M)(\boldsymbol{\phi}^{*}-1)+O(C_{\eta}\sqrt{d/n}\|\boldsymbol{w}\|_{2})\\
 & +O(\|M\|_{2}(\sqrt{k}+|\mathrm{tr}(M)|/\|M\|_{2})\sqrt{d/n})+O[C_{\eta}(\|\boldsymbol{w}\|_{2}+(\sqrt{k}+|\mathrm{tr}(M)|/\|M\|_{2})\|M\|_{2})\sqrt{d/n}]\\
= & \mathcal{D}(\boldsymbol{\kappa}^{*})\boldsymbol{w}+\mathcal{D}(M)(\boldsymbol{\phi}^{*}-1)+O[C_{\eta}(\|\boldsymbol{w}\|_{2}+\|M\|_{2})(\sqrt{k}+|\mathrm{tr}(M)|/\|M\|_{2})\sqrt{d/n}]\\
= & \mathcal{D}(\boldsymbol{\kappa}^{*})\boldsymbol{w}+\mathcal{D}(M)(\boldsymbol{\phi}^{*}-1)+O[\delta(\|\boldsymbol{w}\|_{2}+\|M\|_{2})]\ .
\end{align*}

\end{proof}

\section{Proof of Lemma \ref{lem:error-bound-G_j-G_j_start}}

The next lemma bounds the estimation accuracy of $\boldsymbol{\kappa}^{*},\boldsymbol{\phi}^{*}$
. It directly follows sub-gaussian concentration inequality and union
bound.
\begin{lem}
\label{lem:estimation-accuracy-of-3-4-moments} Given $n$ i.i.d.
sampled $\boldsymbol{x}^{(i)}$, $i\in\{1,\cdots,n\}$. With a probability
at least $1-\eta$,
\begin{align*}
\|\boldsymbol{\kappa}-\boldsymbol{\kappa}^{*}\|_{\infty}\leq & C_{\eta}/\sqrt{n}\\
\|\boldsymbol{\phi}-\boldsymbol{\phi}^{*}\|_{\infty}\leq & C_{\eta}/\sqrt{n}
\end{align*}
 provided $n\geq C_{\eta}d$ .
\end{lem}
Denote $G^{*}$ as $G$ in Eq. (\ref{eq:G}) but computed with $\boldsymbol{\kappa}^{*},\boldsymbol{\phi}^{*}$.
The next lemma bounds $\|G_{j,:}-G_{j,:}^{*}\|_{2}$ for any $j\in\{1,\cdots,d\}$.
\begin{proof}
Denote $\boldsymbol{g}=G_{j}$, $\boldsymbol{g}^{*}=G_{j}^{*}$, $\kappa=\boldsymbol{\kappa}_{j}$,
$\mathbf{\phi=\boldsymbol{\phi}}_{j}$,
\begin{align*}
A= & \left[\begin{array}{cc}
1 & \boldsymbol{\kappa}_{j}\\
\boldsymbol{\kappa}_{j} & \boldsymbol{\phi}_{j}-1
\end{array}\right],\ \boldsymbol{b}=\left[\begin{array}{c}
\boldsymbol{\kappa}_{j}\\
\boldsymbol{\phi}_{j}-3
\end{array}\right]\\
A^{*}= & \left[\begin{array}{cc}
1 & \boldsymbol{\kappa}_{j}^{*}\\
\boldsymbol{\kappa}_{j}^{*} & \boldsymbol{\phi}_{j}^{*}-1
\end{array}\right],\ \boldsymbol{b}^{*}=\left[\begin{array}{c}
\boldsymbol{\kappa}_{j}^{*}\\
\boldsymbol{\phi}_{j}^{*}-3
\end{array}\right]\ .
\end{align*}
Then $\boldsymbol{g}=A^{-1}\boldsymbol{b}$, $\boldsymbol{g}^{*}=A^{*-1}\boldsymbol{b}^{*}$
. Since $\mathbb{P}(\boldsymbol{x})$ is $\tau$-MIP, $\|A^{*-1}\|_{2}\leq1/\tau$
. From Lemma \ref{lem:estimation-accuracy-of-3-4-moments},
\begin{align*}
\|A-A^{*}\|_{2}\leq & C\log(d/\eta)/\sqrt{n}\\
\|\boldsymbol{b}-\boldsymbol{b}^{*}\|_{2}\leq & C\log(d/\eta)/\sqrt{n}\ .
\end{align*}
 Define $\Delta_{A}\triangleq A-A^{*}$, $\Delta_{b}\triangleq\boldsymbol{b}-\boldsymbol{b}^{*}$,
$\Delta_{g}\triangleq\boldsymbol{g}-\boldsymbol{g}^{*}$,
\begin{align*}
 & A\boldsymbol{g}=\boldsymbol{b}\\
\Leftrightarrow & (A^{*}+\Delta_{A})(\boldsymbol{g}^{*}+\Delta_{g})=\boldsymbol{b}^{*}+\Delta_{b}\\
\Leftrightarrow & A^{*}\Delta_{g}+\Delta_{A}\boldsymbol{g}^{*}+\Delta_{A}\Delta_{g}=\Delta_{b}\\
\Leftrightarrow & (A^{*}+\Delta_{A})\Delta_{g}=\Delta_{b}-\Delta_{A}\boldsymbol{g}^{*}\\
\Rightarrow & \|(A^{*}+\Delta_{A})\Delta_{g}\|_{2}=\|\Delta_{b}-\Delta_{A}\boldsymbol{g}^{*}\|_{2}\\
\Rightarrow & \|(A^{*}+\Delta_{A})\Delta_{g}\|_{2}\leq\|\Delta_{b}\|_{2}+\|\Delta_{A}\boldsymbol{g}^{*}\|_{2}\\
\Rightarrow & \|(A^{*}+\Delta_{A})\Delta_{g}\|_{2}\leq C\log(d/\eta)/\sqrt{n}+C\log(d/\eta)/\sqrt{n}\|\boldsymbol{g}^{*}\|_{2}\\
\Rightarrow & \|(A^{*}+\Delta_{A})\Delta_{g}\|_{2}\leq C\log(d/\eta)/\sqrt{n}(1+\|\boldsymbol{g}^{*}\|_{2})\\
\Rightarrow & \|(A^{*}+\Delta_{A})\Delta_{g}\|_{2}\leq C\log(d/\eta)/\sqrt{n}(1+\frac{1}{\tau}\|\boldsymbol{b}^{*}\|_{2})\\
\Rightarrow & \|(A^{*}+\Delta_{A})\Delta_{g}\|_{2}\leq C\log(d/\eta)/\sqrt{n}(1+\frac{1}{\tau}\sqrt{\kappa^{2}+(\phi-3)^{2}})\\
\Rightarrow & [\tau-C\log(d)/\sqrt{n}]\|\Delta_{g}\|_{2}\leq C\log(d/\eta)/\sqrt{n}(1+\frac{1}{\tau}\sqrt{\kappa^{2}+(\phi-3)^{2}})\ .
\end{align*}
 When
\begin{align*}
 & \tau-C\log(d/\eta)/\sqrt{n}\geq\frac{1}{2}\tau\\
\Leftrightarrow & n\geq4C^{2}\log^{2}(d/\eta)/\tau^{2}\ ,
\end{align*}
 we have
\begin{align*}
\|\Delta_{g}\|_{2}\leq & \frac{2C}{\tau\sqrt{n}}\log(d/\eta)(1+\frac{1}{\tau}\sqrt{\kappa^{2}+(\phi-3)^{2}})\ .
\end{align*}
 Since $\Delta_{g}$ is a vector of dimension 2, its $\ell_{2}$-norm
bound also controls its $\ell_{\infty}$-norm bound up to constant.
Choose
\begin{align*}
 & n\geq C_{\eta}\frac{1}{\tau}(1+\frac{1}{\tau}\sqrt{\kappa^{2}+(\phi-3)^{2}})/\delta^{2}\ .
\end{align*}
 We have
\begin{align*}
\|\Delta_{g}\|_{\infty}\leq & \delta\ .
\end{align*}

The proof of $H$ is similar.
\end{proof}

\section{Proof of Lemma \ref{lem:concentration-of-mathcal-M-and-W}}
\begin{proof}
To abbreviate the notation, we omit $\hat{\boldsymbol{y}}^{(t)}-\boldsymbol{y}^{(t)}$
and superscript $t$ in the following proof. Denote $\mathcal{H}^{*}=\mathbb{E}\mathcal{H}$
and the expectation of other operators similarly. By construction
in Algorithm \ref{thm:global-convergence-rate-of-Moment-Estimation-Sequence},
\begin{align*}
\mathcal{M}\triangleq & \mathcal{H}-\frac{1}{2}\mathcal{D}(G_{1}\circ\mathcal{P}^{(1)})-\frac{1}{2}\mathcal{D}(G_{2}\circ\mathcal{P}^{(2)})\\
= & \mathcal{H}^{*}+O[\delta(\alpha_{t-1}+\beta_{t-1})]\\
 & -\frac{1}{2}\mathcal{D}(G_{1}^{*}\circ\mathcal{P}^{*(1)})-\frac{1}{2}\mathcal{D}(G_{2}^{*}\circ\mathcal{P}^{*(2)})\\
 & +O[\|G-G^{*}\|_{\infty}(\|\mathcal{P}^{*(1)}\|_{2}+\|\mathcal{P}^{*(2)}\|_{2})]\\
 & +O[\|G-G^{*}\|_{\infty}\delta(\alpha_{t-1}+\beta_{t-1})]\\
= & M^{(t)}-M^{*}+O[\delta(\alpha_{t-1}+\beta_{t-1})]\\
 & +O[\delta(\|\mathcal{P}^{*(1)}\|_{2}+\|\mathcal{P}^{*(2)}\|_{2})]\\
 & +O[\delta^{2}(\alpha_{t-1}+\beta_{t-1})]\\
= & M^{(t)}-M^{*}+O[\delta(\alpha_{t-1}+\beta_{t-1})]\\
 & +O[\delta(\|\mathcal{P}^{*(1)}\|_{2}+\|\mathcal{P}^{*(2)}\|_{2})]\\
= & M^{(t)}-M^{*}+O[\delta(\alpha_{t-1}+\beta_{t-1})]\\
 & +O[\delta(\alpha_{t-1}\|\boldsymbol{\kappa}^{*}\|_{\infty}+\beta_{t-1})\\
 & +\alpha_{t-1}\|\boldsymbol{\phi}^{*}-1\|_{\infty}+\beta_{t-1}\|\boldsymbol{\kappa}^{*}\|_{\infty}]\\
= & M^{(t)}-M^{*}+O[\delta(\alpha_{t-1}+\beta_{t-1})]+O[\delta p(\alpha_{t-1}+\beta_{t-1})]\\
= & M^{(t)}-M^{*}+O[\delta(p+1)(\alpha_{t-1}+\beta_{t-1})]\ .
\end{align*}
 The above requires
\begin{align*}
n\geq & \max\{C_{\eta}\frac{1}{\tau}(1+\frac{1}{\tau}\sqrt{\kappa^{2}+(\phi-3)^{2}})/\delta^{2},C_{\eta}k^{2}d\}\\
= & \max\{C_{\eta}p(\tau\delta)^{-2},C_{\eta}k^{2}d\}\ .
\end{align*}
 Replace $\delta(p+1)$ with $\delta$, the proof is completed.

To bound $\mathcal{W}^{(t)}(\hat{\boldsymbol{y}}^{(t)}-\boldsymbol{y}^{(t)})$,
similarly we have
\begin{align*}
\mathcal{W}= & G_{1}\circ\mathcal{P}^{(1)}+G_{2}\circ\mathcal{P}^{(2)}\\
= & G_{1}^{*}\circ\mathcal{P}^{*(1)}+G_{2}^{*}\circ\mathcal{P}^{*(2)}\\
 & +O[\|G-G^{*}\|_{\infty}\delta(\alpha_{t-1}+\beta_{t-1})]\\
 & +O[\|G-G^{*}\|_{\infty}(\|\mathcal{P}^{*(1)}\|_{2}+\|\mathcal{P}^{*(2)}\|_{2})]\\
= & \boldsymbol{w}^{(t-1)}-\boldsymbol{w}^{*}+O[\delta^{2}(\alpha_{t-1}+\beta_{t-1})]\\
 & +O[\delta p(\alpha_{t-1}+\beta_{t-1})]\\
= & \boldsymbol{w}^{(t-1)}-\boldsymbol{w}^{*}+O[\delta(p+1)(\alpha_{t-1}+\beta_{t-1})]\ .
\end{align*}
\end{proof}

\section{Proof of Theorem \ref{thm:global-convergence-rate-of-Moment-Estimation-Sequence}
\label{sec:proof-of-theorem-global-convergence}}

Denote $\theta(U,V)$ as the largest canonical angle between the subspaces
spanned by the columns of $U$ and $V$ respectively. We need some
tools from the matrix perturbation analysis \citep{stewart_matrix_1990}.
\begin{lem}
\label{lem:svd-perturbation-bound} Let $U$ be the left/right top-$k$
singular vectors of $M$. Suppose $\hat{M}=M+\mathcal{O}(\xi)$ with
left/right singular vectors $\hat{U}$. Then for any $\epsilon\leq[\sigma_{k}(M)-\sigma_{k+1}(M)]/4$,
\begin{align*}
\sin\theta(U,\hat{U})\leq & 2\epsilon/[\sigma_{k}(M)-\sigma_{k+1}(M)]\ .
\end{align*}
\end{lem}
To commerce our proof, we bound the initialization error. By construction
in Algorithm \ref{alg:moment-estimation-sequence-method} and Lemma
\ref{lem:concentration-of-mathcal-M-and-W}
\begin{align*}
\mathcal{M}^{(0)}(\boldsymbol{y}^{(0)})= & -M^{*}+O(\delta\epsilon_{0})
\end{align*}
 where $\epsilon_{0}=\|\boldsymbol{w}^{*}\|_{2}+\|M^{*}\|_{2}$. Then
from Lemma \ref{lem:svd-perturbation-bound},
\[
\sin\theta(U^{*},U^{(0)})\leq2\delta\epsilon_{0}/\sigma_{k}^{*}\ .
\]

Support at step $t-1$,
\begin{align*}
 & \bar{U}^{(t-1)}{}^{\top}\bar{U}^{(t-1)}=I\\
 & M^{(t-1)}=\bar{U}^{(t-1)}V^{(t-1)}{}^{\top}=U^{(t-1)}\bar{V}^{(t-1)}{}^{\top}\\
 & \epsilon_{t-1}\triangleq\|\boldsymbol{w}^{(t-1)}-\boldsymbol{w}^{*}\|_{2}+\|M^{(t-1)}-M^{*}\|_{2}\\
 & \theta_{t-1}\triangleq\theta(U^{(t-1)},U^{*})\\
 & \alpha_{t-1}\triangleq\tan\theta_{t-1}
\end{align*}
According to Algorithm \ref{alg:moment-estimation-sequence-method},
\begin{align*}
U^{(t)}= & V^{(t-1)}-\mathcal{M}^{(t)}(\hat{\boldsymbol{y}}^{(t)}-\boldsymbol{y}^{(t)})\bar{U}^{(t-1)}\\
= & V^{(t-1)}-[M^{(t-1)}{}^{\top}-M^{*}+O(\delta\epsilon_{t-1})]\bar{U}^{(t-1)}\\
= & M^{(t-1)}{}^{\top}\bar{U}^{(t-1)}-[M^{(t-1)}{}^{\top}-M^{*}+O(\delta\epsilon_{t-1})]\bar{U}^{(t-1)}\\
= & M^{*}\bar{U}^{(t-1)}+O(\delta\epsilon_{t-1})\bar{U}^{(t-1)}\ .
\end{align*}
\[
\bar{U}^{(t)}R^{(t)}=\mathrm{qr}(U^{(t)})
\]
\begin{align*}
V^{(t)}= & V^{(t-1)}\bar{U}^{(t-1)}{}^{\top}\bar{U}^{(t)}-\mathcal{M}^{(t)}(\hat{\boldsymbol{y}}^{(t)}-\boldsymbol{y}^{(t)})\bar{U}^{(t)}\\
= & M^{(t-1)}{}^{\top}\bar{U}^{(t)}-\mathcal{M}^{(t)}(\hat{\boldsymbol{y}}^{(t)}-\boldsymbol{y}^{(t)})\bar{U}^{(t)}\\
= & M^{(t-1)}{}^{\top}\bar{U}^{(t)}-[M^{(t-1)}{}^{\top}-M^{*}+O(\delta\epsilon_{t-1})]\bar{U}^{(t)}\\
= & M^{*}\bar{U}^{(t)}+O(\delta\epsilon_{t-1})\bar{U}^{(t)}\ .
\end{align*}

\begin{align*}
M^{(t)}= & \bar{U}^{(t)}V^{(t)}{}^{\top}\\
= & \bar{U}^{(t)}[M^{*}\bar{U}^{(t)}+O(\delta\epsilon_{t-1})\bar{U}^{(t)}]{}^{\top}\\
= & \bar{U}^{(t)}\bar{U}^{(t)}{}^{\top}M^{*}+\bar{U}^{(t)}\bar{U}^{(t)}{}^{\top}O(\delta\epsilon_{t-1})
\end{align*}

\begin{align*}
\|M^{(t)}-M^{*}\|_{2}= & \|\bar{U}^{(t)}\bar{U}^{(t)}{}^{\top}M^{*}+\bar{U}^{(t)}\bar{U}^{(t)}{}^{\top}O(\delta\epsilon_{t-1})-M^{*}\|_{2}\\
= & \|(\bar{U}^{(t)}\bar{U}^{(t)}{}^{\top}-I)M^{*}\|_{2}+\|\bar{U}^{(t)}\bar{U}^{(t)}{}^{\top}O(\delta\epsilon_{t-1})\|_{2}\\
\leq & \sin\theta(\bar{U}^{(t)},U^{*})\sigma_{1}^{*}+\delta\epsilon_{t-1}\\
\leq & \alpha_{t}\sigma_{1}^{*}+\delta\epsilon_{t-1}\ .\quad(\sin\theta\leq\tan\theta)
\end{align*}

Next we bound $\theta(U^{(t)},U^{*})$.
\begin{align*}
\sin\theta_{t}= & \|U_{\perp}^{*}{}^{\top}U^{(t)}\|_{2}\\
= & \|U_{\perp}^{*}{}^{\top}[M^{*}\bar{U}^{(t-1)}+O(\delta\epsilon_{t-1})\bar{U}^{(t-1)}]\|_{2}\\
\leq & \|U_{\perp}^{*}{}^{\top}M^{*}\bar{U}^{(t-1)}\|_{2}+\delta\epsilon_{t-1}\\
\leq & \delta\epsilon_{t-1}
\end{align*}
\begin{align*}
\cos\theta_{t}= & \sigma_{k}\{U^{*}{}^{\top}U^{(t)}\}\\
= & \sigma_{k}\{U^{*}{}^{\top}[M^{*}\bar{U}^{(t-1)}+O(\delta\epsilon_{t-1})\bar{U}^{(t-1)}]\}\\
\geq & \sigma_{k}\{U^{*}{}^{\top}M^{*}\bar{U}^{(t-1)}\}-\delta\epsilon_{t-1}\\
\geq & \sigma_{k}^{*}\sigma_{k}\{U^{*}{}^{\top}\bar{U}^{(t-1)}\}-\delta\epsilon_{t-1}\\
= & \sigma_{k}^{*}\cos\theta_{t-1}-\delta\epsilon_{t-1}
\end{align*}
\begin{align*}
\tan\theta_{t}= & \frac{\sin\theta_{t}}{\cos\theta_{t}}\leq\frac{\delta\epsilon_{t-1}}{\sigma_{k}^{*}\cos\theta_{t-1}-\delta\epsilon_{t-1}}\ .
\end{align*}
We require
\begin{align*}
 & \cos\theta_{t-1}\geq\frac{1}{\sqrt{5}}\\
\Leftarrow & \cos\theta_{0}\geq\frac{1}{\sqrt{5}}\\
\Leftarrow & \sin\theta_{0}\leq\frac{2}{\sqrt{5}}\\
\Leftarrow & 2\delta\epsilon_{0}/\sigma_{k}^{*}\leq\frac{2}{\sqrt{5}}\\
\Leftarrow & \delta\leq\frac{1}{\sqrt{5}}\frac{\sigma_{k}^{*}}{\epsilon_{0}}\quad(*)
\end{align*}
 and require
\begin{align*}
 & \delta\epsilon_{t-1}\leq\frac{1}{2\sqrt{5}}\sigma_{k}^{*}\\
\Leftarrow & \delta\leq\frac{1}{2\sqrt{5}}\frac{\sigma_{k}^{*}}{\epsilon_{0}}\quad(*)
\end{align*}
 Then
\begin{align*}
\tan\theta_{t}\leq & \frac{2\sqrt{5}}{\sigma_{k}^{*}}\delta\epsilon_{t-1}\ .
\end{align*}

To bound $\|\boldsymbol{w}^{(t)}-\boldsymbol{w}^{*}\|_{2}$ , from
Lemma \ref{lem:concentration-of-mathcal-M-and-W},
\begin{align*}
\boldsymbol{w}^{(t)}= & \boldsymbol{w}^{(t-1)}-\mathcal{W}^{(t)}(\hat{\boldsymbol{y}}^{(t)}-\boldsymbol{y}^{(t)})\\
= & \boldsymbol{w}^{(t-1)}-[\boldsymbol{w}^{(t-1)}-\boldsymbol{w}^{*}+O(\delta\epsilon_{t-1})]\\
= & \boldsymbol{w}^{*}+O(\delta\epsilon_{t-1})\ .
\end{align*}
 Therefore
\begin{align*}
\|\boldsymbol{w}^{(t)}-\boldsymbol{w}^{*}\|_{2}\leq & \delta\epsilon_{t-1}\ .
\end{align*}

To bound $\epsilon_{t}$,
\begin{align*}
\epsilon_{t}= & \|\boldsymbol{w}^{(t)}-\boldsymbol{w}^{*}\|_{2}+\|M^{(t)}-M^{*}\|_{2}\\
\leq & \delta\epsilon_{t-1}+\alpha_{t}\sigma_{1}^{*}+\delta\epsilon_{t-1}\\
\leq & \alpha_{t}\sigma_{1}^{*}+2\delta\epsilon_{t-1}\\
\leq & \frac{2\sqrt{5}}{\sigma_{k}^{*}}\delta\epsilon_{t-1}\sigma_{1}^{*}+2\delta\epsilon_{t-1}\\
\leq & (2\sqrt{5}\sigma_{1}^{*}/\sigma_{k}^{*}+2)\delta\epsilon_{t-1}\ .
\end{align*}
To require that $\epsilon_{t}$ is non-increasing,
\begin{align*}
 & (2\sqrt{5}\sigma_{1}^{*}/\sigma_{k}^{*}+2)\delta\leq1\\
\Leftarrow & \delta\leq\frac{1}{2\sqrt{5}\sigma_{1}^{*}/\sigma_{k}^{*}+2}\ .\quad(*)
\end{align*}

Merge all requirements marked by $(*)$, we require
\begin{align*}
\delta\leq & \min\{\frac{1}{2\sqrt{5}\sigma_{1}^{*}/\sigma_{k}^{*}+2},\frac{1}{2\sqrt{5}}\frac{\sigma_{k}^{*}}{\epsilon_{0}},\frac{1}{\sqrt{5}}\frac{\sigma_{k}^{*}}{\epsilon_{0}}\}\ .
\end{align*}
 And the convergence rate is
\begin{align*}
\epsilon_{t}\leq & (2\sqrt{5}\sigma_{1}^{*}/\sigma_{k}^{*}+2)\delta\epsilon_{t-1}\\
\leq & [(2\sqrt{5}\sigma_{1}^{*}/\sigma_{k}^{*}+2)\delta]^{t}\epsilon_{0}\ .
\end{align*}

\end{document}